\pdfoutput=1

\documentclass[11pt]{article}

\usepackage[final]{acl}

\usepackage{times}
\usepackage{latexsym}

\usepackage[T1]{fontenc}

\usepackage[utf8]{inputenc}

\usepackage{microtype}

\usepackage{inconsolata}

\usepackage[utf8]{inputenc}
\usepackage{array}
\usepackage[linesnumbered,ruled,vlined]{algorithm2e}
\usepackage{algorithmic}
\usepackage{amsmath}
\usepackage{amsfonts}
\usepackage{amssymb}
\usepackage{multirow}
\usepackage{booktabs}
\usepackage{url}
\usepackage{enumitem}
\usepackage{epstopdf}
\usepackage{makecell}
\usepackage{graphicx}
\usepackage{appendix}
\usepackage{bm}
\usepackage{soul,color}
\usepackage{adjustbox}
\usepackage[normalem]{ulem}
\usepackage{balance}
\usepackage{multicol}

\newcommand{\method}{\textsc{ProLINK}\xspace} 

\newcommand{\kaiupdate}[1]{\textcolor{black}{#1}}

\newcommand{\graph}{\mathcal{G}}
\newcommand{\gtrain}{\mathcal{G}_{\textit{tr}}}
\newcommand{\ginf}{\mathcal{G}_{\textit{inf}}}

\newcommand{\etrain}{\mathcal{E}_{\textit{tr}}}
\newcommand{\einf}{\mathcal{E}_{\textit{inf}}}
\newcommand{\rtrain}{\mathcal{R}_{\textit{tr}}}
\newcommand{\rinf}{\mathcal{R}_{\textit{inf}}}
\newcommand{\ttrain}{\mathcal{T}_{\textit{tr}}}
\newcommand{\tinf}{\mathcal{T}_{\textit{inf}}}

\newcommand{\ents}{\mathcal{E}}
\newcommand{\rels}{\mathcal{R}}
\newcommand{\triples}{\mathcal{T}}

\title{LLM as Prompter: Low-resource Inductive Reasoning on Arbitrary Knowledge Graphs}

\author{
Kai Wang$^{\dag}$, Yuwei Xu$^{\dag}$, Zhiyong Wu$^{\ddag}$, Siqiang Luo$^\dag$\thanks{~~The Corresponding Author}\\
 $^{\dag}$Nanyang Technological University 
 $^{\ddag}$Shanghai Artificial Intelligence Laboratory \\
 $^{\dag}$kai\_wang@ntu.edu.sg, S220060@e.ntu.edu.sg, siqiang.luo@ntu.edu.sg\\
 $^{\ddag}$wuzhiyong@pjlab.org.cn\\
}

\begin{document}
\maketitle
\begin{abstract}
   Knowledge Graph (KG) inductive reasoning, which aims to infer missing facts from new KGs that are not seen during training, has been widely adopted in various applications. One critical challenge of KG inductive reasoning is handling low-resource scenarios with scarcity in both textual and structural aspects. In this paper, we attempt to address this challenge with Large Language Models (LLMs).
   Particularly, we utilize the state-of-the-art LLMs {to generate a graph-structural prompt}
  to enhance the pre-trained Graph Neural Networks (GNNs), 
  which brings us new methodological insights into the KG inductive reasoning methods, as well as high generalizability in practice. 
  On the methodological side, we introduce a novel pretraining and prompting framework \method, designed for low-resource inductive reasoning across arbitrary KGs without requiring additional training. 
  On the practical side, we experimentally evaluate our approach on 36 low-resource KG datasets and find that \method outperforms previous methods in three-shot, one-shot, and zero-shot reasoning tasks, exhibiting {average} performance improvements by 20\%, 45\%, and 147\%, respectively.
  \kaiupdate{Furthermore, \method demonstrates strong robustness for various LLM promptings as well as full-shot scenarios.}
  Our source code is available on \url{https://github.com/KyneWang/ProLINK}.
\end{abstract}

\maketitle
\section{{Introduction}}

\begin{figure}[!tb]
\centering
\includegraphics[width=0.48\textwidth]{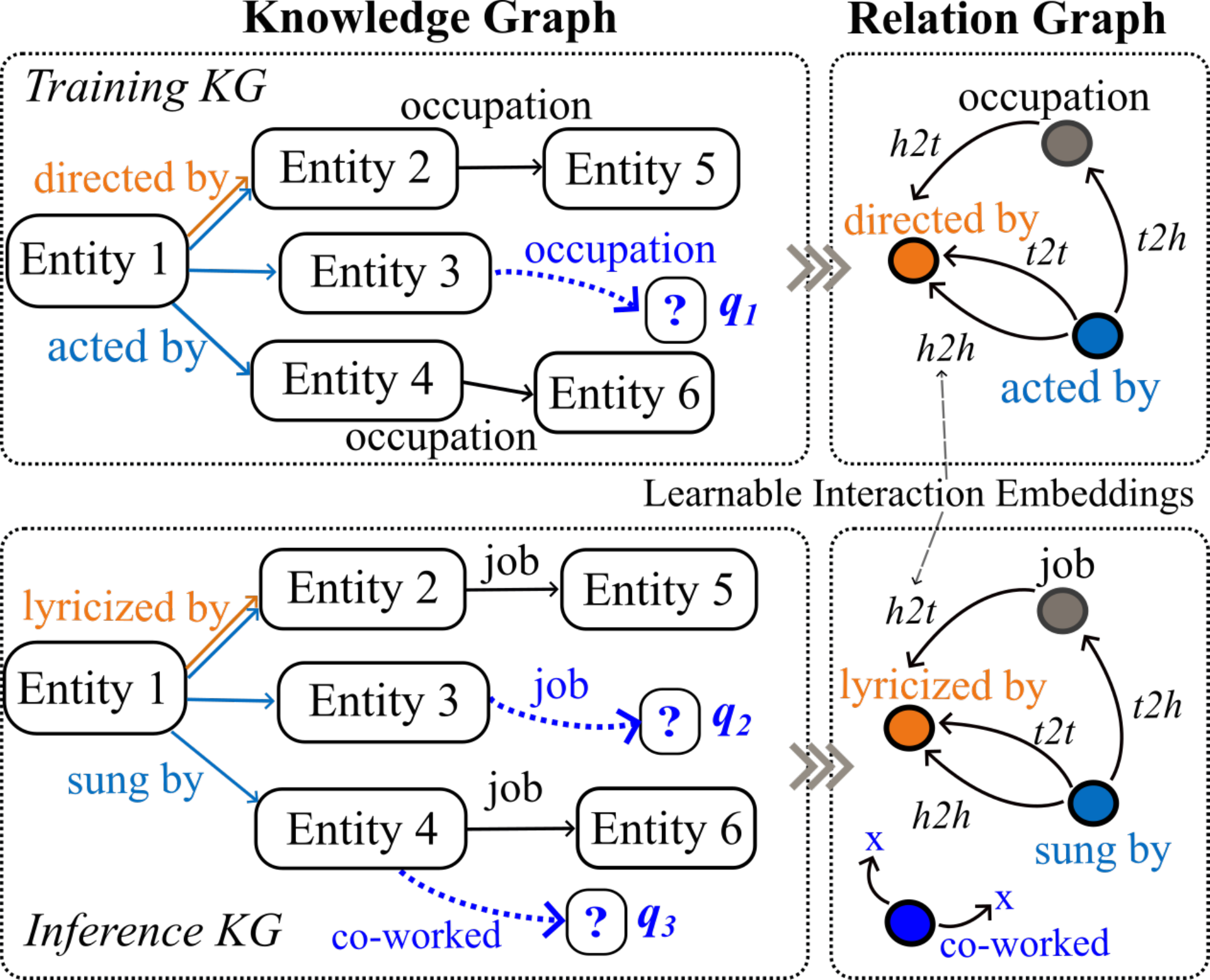}
\vspace{-5mm}
\caption{Illustrations of knowledge graphs and corresponding relation graphs.
The interaction edge $h2t$ means the source relation has a head entity in the KG which is a tail entity of the target relation.
}
\vspace{-8mm}
\label{fig:0}
\end{figure}

Knowledge Graph (KG) Reasoning, also known as KG Link Prediction, aims at inferring new facts from existing KGs
in the triple format (head entity, relation, tail entity)\cite{OurIJCAI, OurRotL, ACL23-KGR2}. This technique has been widely studied and applied in various domains, including information retrieval, e-commerce recommendations, drug discovery, and financial prediction~\cite{Finan1, Drug1, Drug2, ACL23-KGR3}. 
For example, the query ($q_1$) in Figure \ref{fig:0} consists of `{\it Entity 3}' and a relation type `{\it occupation}'. Its expected answer for KG reasoning is `{\it Entity 6}'.



The dynamic nature of real-world KGs fosters recent research interest in {\it inductive KG reasoning}—inferring from new KGs which have entities/relations unseen during training~\cite{DRUM-NIPS19,GraIL-ICML19}.
In contrast to earlier methods that learn graph-specific embeddings for the training KG \cite{TransE,DistMult-paper,RotatE}, recent inductive reasoning models are trained to address the query ($q_1$) in the training KG, enabling them to answer the query ($q_2$) in a fully new KG \cite{ingram,ultra}. 
As illustrated in Figure \ref{fig:0}, this transferability stems from the shared interaction patterns in relation graphs, which can be captured by Graph Neural Networks (GNNs) \cite{GCN-ICLR17,lab_4_SCARA,lab_5_SpikingGCN,lab_10_SIMGA} without using entity/relation textual information.

One critical challenge of KG inductive reasoning is \textit{handling low-resource scenarios marked by scarcity of both textual and structural information}.
For the query ($q_3$) in Figure \ref{fig:0}, the inference KG lacks textual context for entities and has no support triple for the `{\it co-worked}' relation.
Such low-resource scenarios frequently occur in specialized or emerging domains due to long-tail distribution and new relation additions \cite{FewKGR-SIGKDD23,FewKGR-ICLR23}, hindering the wide adoption of KG inductive reasoning.
Text-based methods, even the current powerful Large Language Models (LLMs) \cite{Llama2,Gpt4}, are constrained by limited textual data and complicated graph structures, while graph-based methods struggle with few-shot relation types that lack enough interaction edges in the relation graph\footnote{We verified this challenge in preliminaries in Section \ref{sec:2.3}.}.

Human reasoning, however, may address the above low-resource queries without expert knowledge or prior learning. 
Simply leveraging limited relation semantics to understand graph structures, humans can successfully deduce that the `{\it co-worked}' relation should connect two persons, suggesting the answer entity can be a head entity of the `{\it job}' relation and near the query entity `{\it Entity 4}'.
Inspired by the human reasoning process which mostly relies on relation semantics, 
we pose a question: 
{
\textit{Can LLMs be used to emulate human reasoning for relation semantic understanding,
thereby enhancing the GNN-based inductive reasoning?}}


This work gives a positive answer to the above question and presents an approach that leverages LLM's basic language power to improve KG inductive reasoning in any low-resource scenarios without additional model training for new KGs.
This ability is crucial for elevating the generalizability of AI technologies to handle data dynamics in the real world. 
Specifically, we propose a novel pretraining and \textbf{Pro}mpting framework \method for \textbf{L}ow-resource \textbf{IN}ductive reasoning across arbitrary \textbf{K}Gs.
First, \method pretrains a GNN-based KG reasoner with novel techniques enhancing few-shot prediction performance.
Then, given a new inference KG with sparse relation types, \method employs a pre-trained LLM to construct a prompt graph through concise relation descriptions (a dozen words per relation).
The prompt graph is calibrated to eliminate noise and subsequently injected into the relation graph of the KG to improve the performance of the GNN reasoner. 
In summary, this work has four novel contributions:
\begin{itemize}[noitemsep,topsep=0pt,parsep=0pt,partopsep=0pt,leftmargin=10pt]
    \item We introduce a new KG reasoning problem, i.e., low-resource inductive reasoning on KGs with arbitrary entity and relation vocabularies, which generalizes most KG link prediction tasks.
    \item To the best of our knowledge, this is the first work leveraging LLMs as the graph prompter for inductive KG reasoning, potentially inspiring further innovation in the research community.
    \item We design a unique {\it pretraining and prompting} framework \method, containing several novel techniques for low-resource inductive reasoning and prompt graph generation.
    \item We construct 36 low-resource inductive datasets from real-world KGs, in which \method outperforms previous state-of-the-art methods in both few-shot and zero-shot reasoning tasks.
\end{itemize}
\section{Background}
\label{sec:2}
\subsection{Notations and Definitions}


Let $\graph = \{\ents, \rels, \triples\}$ denote a Knowledge Graph (KG), where $\ents, \rels$ are the sets of entities and relations. $\triples=\{(e_h,r,e_t)|e_h,e_t \in \ents,r \in \rels\}$ is the set of factual triples of the KG, where $e_h$, $r$ and $e_t$ are called a triple's head entity, relation and tail entity, respectively.
Given a query $(e_q, r_q)$ containing a query entity $e_q\in \ents$ and a query relation $r_q\in \rels$, the KG reasoning task aims to identify the correct entity $e_a \in \ents$, such that the triple $(e_q, r_q, e_a)$ or $(e_a, r_q, e_q)$ is a valid factual triple of $\graph$.


\noindent\textbf{KG Inductive Reasoning:}
Given a model trained on a knowledge graph $\gtrain = \{\etrain, \rtrain, \ttrain\}$, 
the KG reasoning task in the fully inductive scenario evaluates the trained model on a new inference graph $\ginf = \{\einf, \rinf, \tinf\}$, in which all entities and relations are different from those in $\gtrain$, i.e., $\einf \cap \etrain = \emptyset$ and $\rinf \cap \rtrain = \emptyset$.

\noindent\textbf{K-shot Inductive Reasoning:} In the inductive scenario with $\ginf = \{\einf, \rinf, \tinf\}$, given a new relation type $r_q \notin \rinf$ and support triples $\triples_{r_q} = \{(e_h,r_q,e_t)|e_h,e_t\in\einf\}(|\triples_{r_q}|=K)$,
the $K$-shot reasoning task is to predict over a query set $\{(e_q, r_q)|e_q\in\einf\}$ with the augmented KG $\ginf^{r_q} = \{\einf, \rinf\cup\{r_q\}, \tinf\cup\triples_{r_q}\}$. 

In this work, we focus on the low-resource challenge of KG inductive reasoning, inferring not merely unseen but also few-shot relation types. 
Existing methods for few-shot link prediction \cite{FewKGR-EMNLP19,FewKGR-AAAI20,FewKGR-NIPS22,FewKGR-ICLR23} either only work on the training graph structure or have to use graph-specific textual or ontological information to train, therefore cannot achieve inductive reasoning on arbitrary KGs. 
Other related studies also have difficulty accomplishing this low-resource task, including entity-level inductive reasoning \cite{DRUM-NIPS19,GraIL-ICML19,NBF-NIPS21,REDGNN-WWW22} and text-based inductive reasoning \cite{TextKGR-BLP-WWW21, TextKGR-KEPLER-TACL21, TextKGR-StATIK-NAACL21,TextKGR-RAILD-IJCKG22}. Detailed related work is introduced in the Appendix \ref{sec:relatedwork}.

\subsection{Baseline: ULTRA}
\label{sec:2.2}
We first introduce ULTRA \cite{ultra}, the state-of-the-art GNN-based model for inductive reasoning on entirely new KGs.
Its core idea is leveraging the `invariance' of the KG relational structure. With the nature of the triple form, there are four basic interaction types when two relations are connected, i.e., \emph{tail-to-head (t2h)}, \emph{head-to-head (h2h)}, \emph{head-to-tail (h2t)}, and \emph{tail-to-tail (t2t)}. As shown in Figure \ref{fig:0}, different KGs may have similar patterns in the relation interactions.
Thereby, the pre-trained embeddings of four interaction types can be universally shared across KGs to parameterize any unseen relations.

Given an inference KG $\graph=\{\ents,\rels,\triples\}$, ULTRA constructs a relation graph $\graph_{\it r}=\{\rels, \mathcal{R}_{\it fund}, \triples_{\it r}\}$ from the original triple data $\triples$, in which each node is a relation type and edges have four interaction types $\mathcal{R}_{\it fund}$. After adding inverse relations into $\graph$ \cite{NBF-NIPS21,REDGNN-WWW22}, $\graph_{\it r}$ would contain $2|\rels|$ nodes. 
Then, ULTRA employs a graph neural network ${\it GNN}_r(\cdot)$ over the relation graph $\graph_{\it r}$, and obtains the relative representation of each relation conditioned on a query, which then can be used by any off-the-shelf entity-level GNN-based models ${\it GNN}_e(\cdot)$ for KG reasoning \cite{NBF-NIPS21,REDGNN-WWW22,OurGraPE}.
Specifically, given a query $q=(e_q, r_q)$, the score $p(e_q, r_q, e_t)$ of one candidate entity $e_t$ is calculated as follows:
\begin{align}
&\mathbf{R}_q={\it GNN}_r(\theta_r, \mathbf{r}_q, \graph_{\it r}) \label{ultra:gnnr}, \\
&\mathbf{E}_q={\it GNN}_e(\theta_e, \mathbf{e}_q, \mathbf{R}_q, \graph) \label{ultra:gnne}, \\
&p(e_q, r_q, e_t) = {\it MLP}(\mathbf{E}_q[e_t]).
\label{ultra:mlp}
\end{align}
where $\theta_r,\theta_e$ denote the parameters of two GNN modules, and $\mathbf{e}_q,\mathbf{r}_q$ are initialized embedding vectors of $e_q$ and $r_q$\footnote{In the implementation, they are initialized as all-one vectors whereas other nodes in the graph are initialized with zeros, which is verified generalizing better to unseen graphs.}.
Because ULTRA does not require any input features of entities or relations nor learn graph-specific entity or relation embeddings, it enables inductive reasoning across arbitrary KGs.

\begin{figure}[!tb]
\centering
\includegraphics[width=0.48\textwidth]{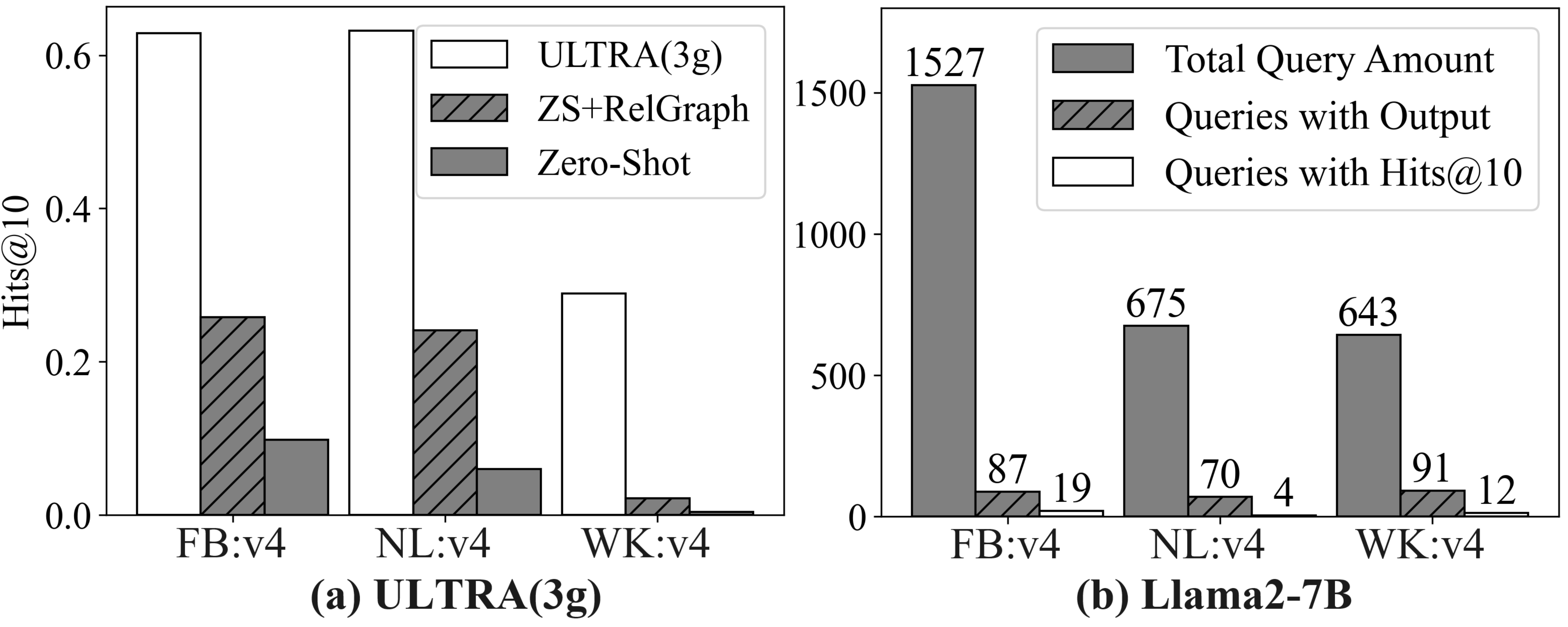}
\vspace{-8mm}
\caption{\kaiupdate{Preliminary results on ULTRA and Llama2.}}
\label{fig:r1}
\vspace{-6mm}
\end{figure} 

\subsection{Preliminaries: Low-resource Challenge}
\label{sec:2.3}
To explore the low-resource challenge, we verify the performance of the GNN-based ULTRA \cite{ultra} and the LLM-based Llama2 \cite{Llama2} in zero-shot inductive reasoning.
On three real-world KG benchmarks, the zero-shot condition is created by removing all $r_q$-involved triples in $\tinf$ for each query relation $r_q$.
The preliminary results of the two pre-trained models are shown in Figure \ref{fig:r1}.

In Figure \ref{fig:r1}(a), we observe that the zero-shot performance of ULTRA(3g) drops sharply compared to its original version.
It indicates the inductive ability of ULTRA highly relies on sufficient support triples in the inference graph. 
Surprisingly, ULTRA obtains better performance after adding the complete relation graph built on the original $\ginf$, which motivates us to enhance the relation graph in low-resource scenarios.
Following recent work \cite{InstructGLM}, we further evaluate the zero-shot graph reasoning performance of Llama2-7B by converting queries and KG subgraphs into textual questions\footnote{For a fair comparison, both our methods and here use only short relation context instead of entity context.}.
Due to the context window limitations of Llama2, we select hundreds of queries whose answers are in the 2-hop neighborhood subgraph of the query entity. Even though, the results are not promising. In Figure \ref{fig:r1}(b), aside from issues related to excessive length or incorrect formatting, only dozens of requests obtain standard outputs and even fewer include the correct answer in the top ten entities outputted (Hits@10).

In summary, the above preliminary results expose the challenges faced by existing methods in low-resource scenarios for KG inductive reasoning, highlighting the necessity for innovative solutions to more effectively utilize sparse data.

\vspace{-2mm}
\section{Methodology}
\label{sec:3}

\begin{figure}[!tb]
\centering
\includegraphics[width=0.47\textwidth]{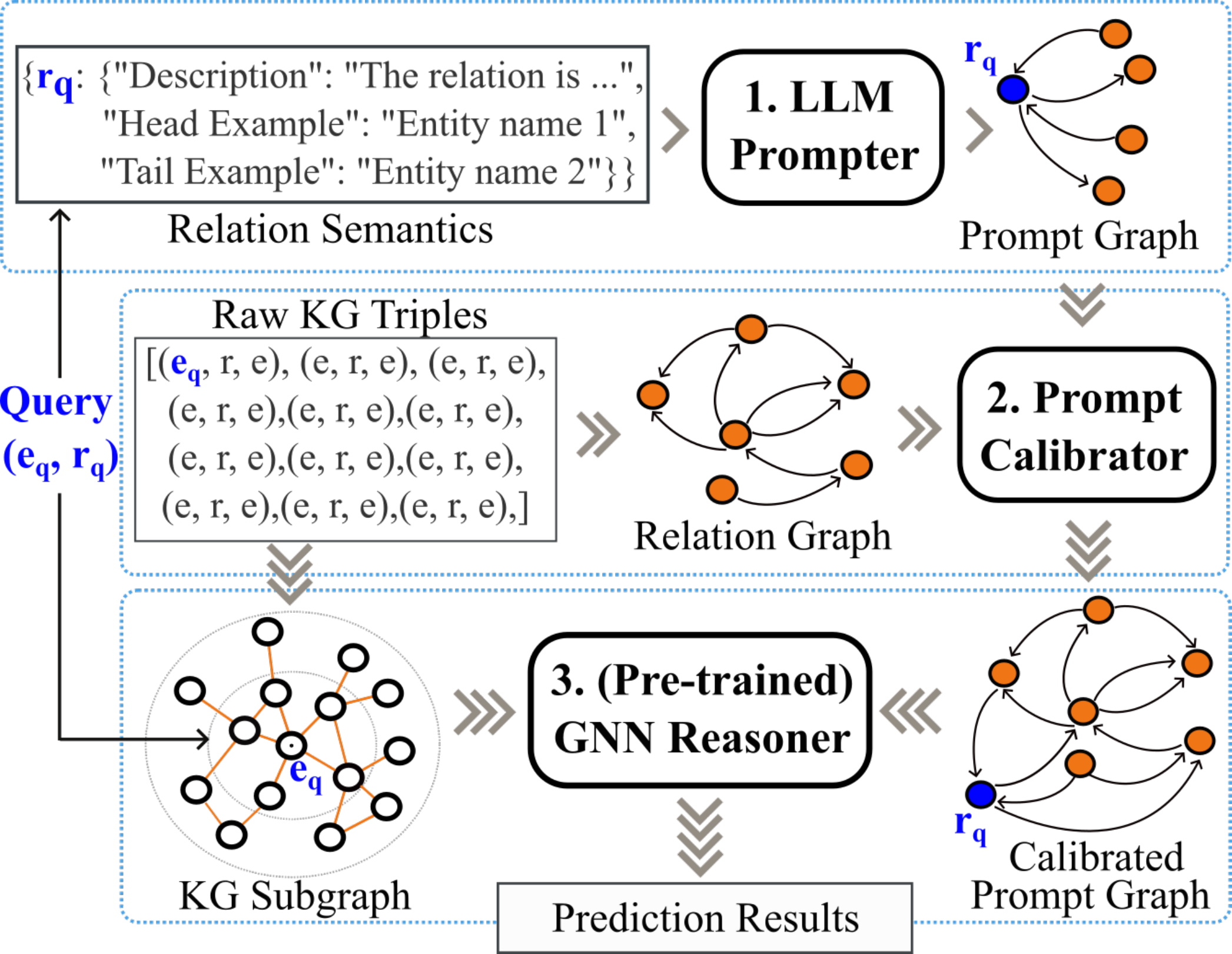}
\vspace{-2mm}
\caption{\method prompting process.}
\label{fig:1}
\vspace{-4mm}
\end{figure} 


This work presents a novel pre-training and prompting framework, \method, which employs pre-trained language models (LLMs) and graph neural networks (GNNs) to handle semantic and structural information, respectively. 
As illustrated in Figure~\ref{fig:1}, 
for a few-shot query relation in a new inferring KG, 
the pre-trained {\it GNN Reasoner} (Section \ref{sec:3.1}) infers from KG subgraphs without model fine-tuning, guided by a relation-specific prompt graph. 
The prompt graph is constructed by a frozen {\it LLM Prompter} (Section \ref{sec:3.2}) from relation semantics, and then calibrated by {\it Prompt Calibrator} (Section \ref{sec:3.3}) to mitigate noise.

\vspace{-2mm}
\subsection{GNN-based Reasoner}
\label{sec:3.1}
Our GNN-based reasoner follows the basic framework of ULTRA \cite{ultra} in Section \ref{sec:2.2}. 
Given a $\graph=\{\ents,\rels,\triples\}$, we first construct the relation graph $\graph_{\it r}=\{\rels, \mathcal{R}_{\it fund}, \triples_{\it r}\}$, which is efficiently obtained from the original graph $\graph$ with sparse matrix multiplications. 
After that, given a query $q=(e_q, r_q)$, we generate relative relation/entity representations $\mathbf{R}_q$/$\mathbf{E}_q$ from the graph structure of $\graph_{\it r}$ and $\graph$, respectively. Then, calculating the triple score $p(e_q, r_q, e_t)$ of any candidate entity $e_t$ follows Equation \ref{ultra:mlp}. 
Specifically, the GNN architecture follows NBFNet~\citep{NBF-NIPS21} with a non-parametric DistMult~\citep{DistMult-paper} message function and sum aggregation.
The relation encoder ${\it GNN}_r(\cdot)$ utilizes randomly initialized edge embeddings for $\mathcal{R}_{\it fund}$. In contrast, ${\it GNN}_e(\cdot)$ initializes the embeddings of edge types using the relative relation embeddings $\mathbf{R}_q$. We suggest consulting the ULTRA paper \cite{ultra} for further details.
To improve the pre-training performance on low-resource inductive reasoning, we propose two enhancing techniques: 

\textbf{Role-aware Relation Encoding:}
The embeddings generated by ${\it GNN}_r(\cdot)$ for relative relations are currently missing vital information: 
the specific role that each relation type assumes in the reasoning process of a query. 
We delineate three unique roles: {\it query relation}, {\it inverse query relation}, and {\it other relation}. 
As a relation may serve different roles across various queries, its embedding vector should reflect the nuances of its designated role.
To imbue this role-aware capability, we introduce trainable role embeddings $\mathbf{R}_o \in \mathbb{R}^{[3 \times d]}$ to augment each relation embedding via a two-layer MLP, formulated as:
$\hat{\mathbf{r}} = \delta\Big(\bm{W}_{2} \delta\big(\bm{W}_{1} (\mathbf{R}_q[r]:\mathbf{R}_o[{\it role}_q(r)])\big)\Big)$,
where $\mathbf{R}_q[r]$ is the relative relation vector of $r$, concatenated with its specific role vector as determined by ${\it role}_q(\cdot)$.
The enhanced relation embeddings $\hat{\mathbf{R}}_q$ are then utilized in Equation \ref{ultra:gnne} replacing $\mathbf{R}_q$.
This component requires far fewer parameters than the original ULTRA. Therefore, efficiency issues can usually be ignored.

\textbf{Low-resource Pretraining Objective:}
The original ULTRA is trained by minimizing the binary cross entropy loss over positive and negative triples. 
Due to the sufficient support triples in the training KG, the GNN reasoner would prioritize reasoning patterns related to these triples, which would not present in low-resource scenarios.
To prompt GNNs to derive insights from a broader range of relations,
we devise an extra pre-training task that operates on 'pseudo' low-resource KGs. 
Specifically, for queries whose relation is $r_q$, we construct a specific KG $\gtrain^{r_q}$, where support triples containing $r_q$ or its inverse relation are randomly masked, using a hyperparameter $\gamma$ to determine the masking proportion. 
The total loss pretraining on the original KG and masked KGs is calculated as follows:
\vspace{-2mm}
\begin{align}
   \small
   \begin{split}
    &\mathcal{L}_{\graph} = - \log p_{\graph}(e_q, r_q, e_a) - \frac{1}{n}\sum_{i=1}^{n}  \log (1 - p_{\graph}(e_q, r_q, e_i))\nonumber
    \end{split}
    \\
    \small
    \begin{split}
    &\mathcal{L} = \alpha\mathcal{L}_{\gtrain} + (1-\alpha)\mathcal{L}_{\gtrain^{r_q}},
    \end{split}
    \label{eqn:training_loss}
\end{align}
where $p_{\graph}(e_q, r_q, e_a)$ represents the score of a positive triple within the specific KG $\graph$. The set $\{(e_q, r_q, e_i)\}^n_{i=1}$ comprises negative samples, which are generated by corrupting either the head or the tail entity in the positive triple. The hyperparameter $\alpha$ balances two parts of loss.
In practical scenarios, 
we manage two pre-training tasks through controlled batch sampling. With a probability of $(1\text{-}\alpha)$, we sample batches with identical query relations for low-resource pretraining, and otherwise, we sample regular batches for pretraining on the original KG.

\vspace{-2mm}
\subsection{LLM-based Prompter}
\label{sec:3.2}
To improve the GNN reasoner's accuracy in low-resource scenarios,
we create a prompt graph $\graph_{\it p}$ connecting few-shot relations with others according to semantic features, thereby filling in the gaps of the topological relation graph $\graph_{\it r}$. 
This is achieved by a frozen Large Language Model (LLM) extracting relation semantics from concise textual information.
This graph prompting process requires no model fine-tuning, preserving the generalizability across distinct KGs.

Initial trials showed that asking the LLM for all possible relational interactions was inefficient and led to inaccuracies, involving too many requests and often resulting in flawed outputs.
To overcome this, we streamlined the process by using the LLM to determine potential entity types for the heads and tails of relations.
When two relations have matching entity types at their head sides, we assume an {\it h2h} interaction between them (similarly for other interaction types). This strategy cuts down LLM queries to a single one for each relation, simplifying the task and enhancing accuracy.

\begin{figure}[!tb]
\centering
\includegraphics[width=0.47\textwidth]{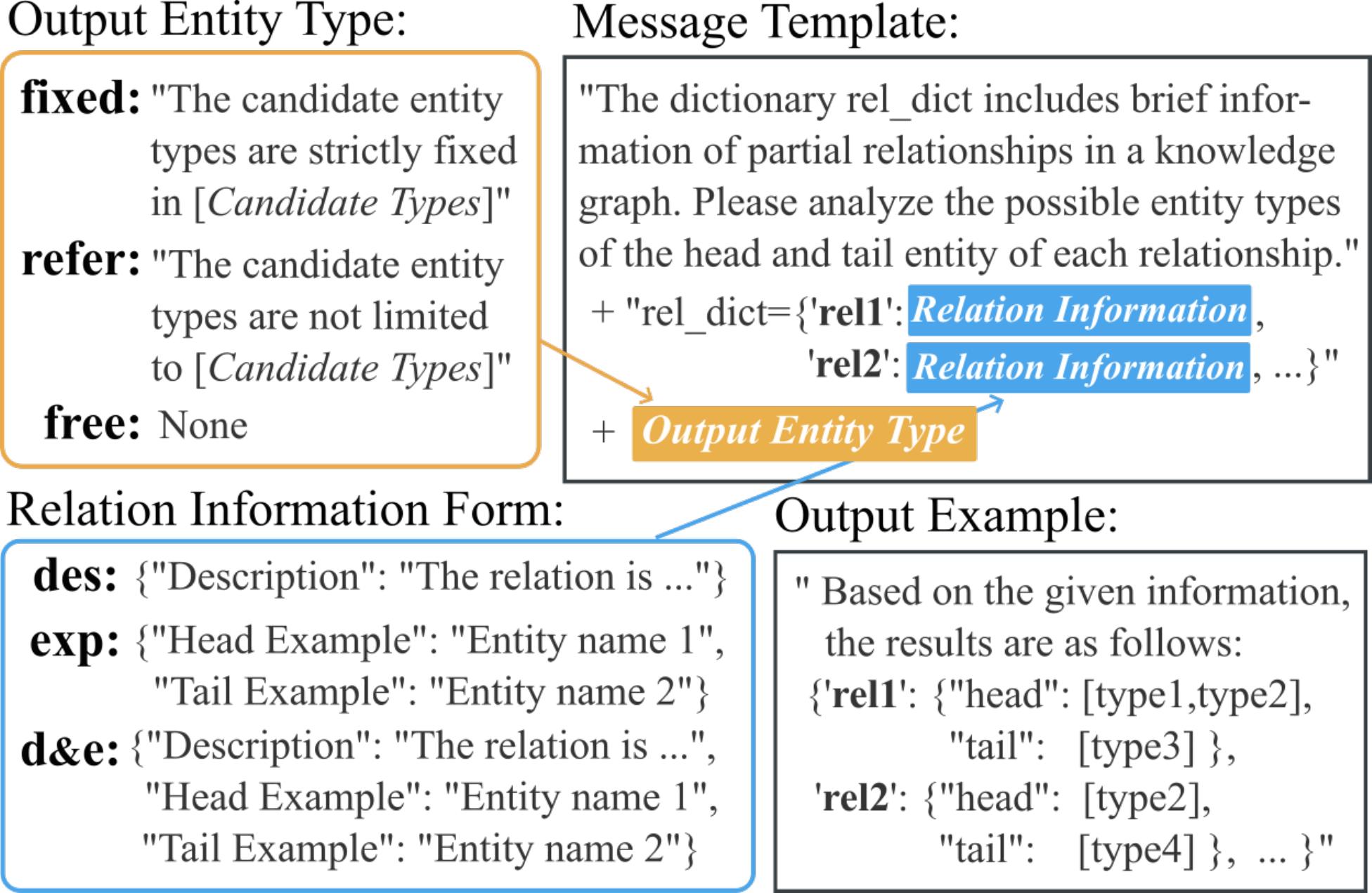}
\vspace{-3mm}
\caption{Examples of instruction prompts for LLMs.}
\label{fig:3}
\vspace{-5mm}
\end{figure}

\textbf{Instruction Prompt Design:}
To obtain entity types of two sides per relation, we design a series of instruction prompts, ensuring detailed guidance for each query relation.
The prompt template is defined as \(\mathcal{P}(\cdot)\), and \(\mathcal{I}=\mathcal{P}(\mathcal{D}_{\rels_s}, L_{\text{et}})\) is the input message to the LLM. \(\rels_s\) is the set of query relations with relation information \(\mathcal{D}_{\rels_s}\). The list of candidate entity types \(L_{\text{et}}\) is utilized to control the range of LLM responses. 
As shown in Figure \ref{fig:3}, \kaiupdate{these prompts are distinct in two aspects:}

\noindent\textit{(1) Relation Information Form:} 
\begin{itemize}[noitemsep,topsep=0pt,parsep=0pt,partopsep=2pt,leftmargin=15pt]
\item \textbf{des:} Short textual description of one relation.
\item \textbf{exp:} Textual entity names of one support triple.
\item \textbf{d\&e:} Both description and example.
\end{itemize}
\textit{(2) Output Entity Type:} 
\begin{itemize}[noitemsep,topsep=0pt,parsep=0pt,partopsep=0pt,leftmargin=15pt]
\item \textbf{fixed:} Limited to predefined entity types.
\item \textbf{refer:} Allow new types besides predefined. 
\item \textbf{free:} No constraints on the type range.
\end{itemize}
As prompt examples in Table \ref{table:proexp} in the Appendix, the semantic information required for each relation is concise, facilitating user editing or automatic generation. 
The list of candidate entity types is domain-dependent; for a general KG, it includes types like person, location, and event. When the type category is \textbf{free}, LLM would output any reasonable entity types with no constraints.

\textbf{Prompt Graph Construct:}
Collecting responses from the LLM,
we obtain the set of entity types $\mathcal{S}(r, \text{`h'})$ and $\mathcal{S}(r, \text{`t'})$ for the head and tail sides of each relation $r$.
Then, we construct the prompt graph $\graph_{\it p}=\{\rels, \mathcal{R}_{\it fund}, \triples_{\it p}\}$, where the relation interactions in $\triples_{\it p}$ obeys the following rules:
\begin{align}
\small
\begin{split}
&\mathcal{S}(r_1, s_1) \cap \mathcal{S}(r_2, s_2) \neq \emptyset \implies \triples_{\it p}[r_1, r_2, s_1, s_2]=1;\nonumber
\end{split}
\\
\small
\begin{split}
&\triples_{\it p}[r_1, r_2, s_1, s_2]=\triples_{\it p}[r_1, r'_2, s_1, s'_2] = \triples_{\it p}[r'_1, r_2, s'_1, s_2];\nonumber
\end{split}
\\
\small
\begin{split}
&\triples_{\it p}[r_1, r_2, s_1, s_2] = \{
\begin{array}{llllr}
\triples_{\it p}[r_2, r_1, s_1, s_2] & s_1 = s_2\\
\triples_{\it p}[r_2, r_1, s'_1, s'_2] & s_1 \neq s_2
\end{array};\nonumber
\end{split}
\label{eq:delta}
\end{align}
where \kaiupdate{$r_1,r_2 \in \rels$ and the side symbol $s_i \in \{\text{`h'}, \text{`t'}\}$}. The inverse relation of $r_i$ is denoted as $r'_i$, and $s'_i$ denotes the opposite side of $s_i$. The connection value $\triples_{\it p}[r_1, r_2, \text{`h'}, \text{`t'}]$ is equal to one when there is an \textit{h2t} edge between $r_1$ and $r_2$, otherwise zero.
The first rule specifies a sufficient condition for establishing relation interactions via entity types, while the other two rules detail the equivalence of inverse relations and inverse interactions. These rules collectively serve to minimize redundant computations in constructing $\graph_{\it p}$. 

\vspace{-2mm}
\subsection{Prompt Calibrator}
\label{sec:3.3}
This component aims to improve the quality of the LLM-based prompt graph $\graph_{\it p}$ leveraging the ground-true information in the topological relation graph $\graph_{\it r}$.
Due to the natural gap between relation semantics and graph-specific topology, $\graph_{\it p}$ cannot cover all expected interactions inevitably.
Besides, mistaken edges would be included due to the relatively loose construction rules and the uncertainty of LLM response quality.
Therefore, we design a novel calibrating process
to extract high-confidence prompting edges that link the query relation with other relations in the KG. As shown in Algorithm \ref{alg:sligen}, 
for each few-shot relation type $r_q$, two learning-free mechanisms are utilized to extract a series of calibrated interaction edges. When inferring $r_q$-involved queries, these interaction edges would be injected into $\graph_{\it r}$ to form the final $r_q$-specific prompt graph $\hat{\graph}_{\it r}^{r_q}$. Detailed descriptions and several examples are introduced in Appendix \ref{app:ddpc}.

\textbf{Few-shot Support Expanding:}
In few-shot scenarios, support triples of this query relation in $\graph_{\it r}$ are valuable.
Therefore, we leverage these triples to find more potential interaction edges via the following rules:
\begin{align}
\small
\begin{split}
\exists r_j,&(r_q, i_r, r_j) \in \graph_{\it r}, i_{r}=[s_1,s_2] \implies \\
&\ \ \ \ \ \ \ \ \ \ \ \ \ \mathcal{S}(r_q, s_1)\leftarrow \mathcal{S}(r_q, s_1) \cup \mathcal{S}(r_j, s_2)\nonumber
\end{split}
\\
\small
\begin{split}
\exists r_k,&(r_q, i_{r1}, r_j),(r_j, i_{r2}, r_k) \in \graph_{\it r}, i_{r1}=[s_1,s_2], \\
&\ \ \ \ \ \ \ \ \ \ i_{r2}=[s_2,s_2] \implies \triples_{\it p}[r_q, r_k, s_1, s_2]=1;\nonumber
\end{split}
\end{align}
where $i_{r}=[s_1,s_2]$ denotes an interaction edge whose type is `$s_1$-to-$s_2$' (e.g., `{\it h2t}'). 
Given the relations involved by support triples, the first rule inserts their entity types into the type set of $r_q$, thereby expanding more potential edges. The second rule adds new interaction edges directly by finding the 2-hop neighbors of the query relation in $\graph_{\it r}$.
These new edges $\triples_{\it ex}^{r_q}$ as well as original $r_q$ related edges $\triples_{\it p}^{r_q}$ will be filtered in the next step.

\begin{algorithm}[!t]
\caption{Prompt Graph Calibrating}
\label{alg:sligen}
\SetKwInOut{Input}{Input}
\SetKwInOut{Output}{Output}
\Input{relation graph $\graph_{\it r}$, prompt graph $\graph_{\it p}$, query relation $r_q$, threshold $\beta$.}
\Output{calibrated prompt graph $\hat{\graph_{\it r}}$.}

Gather $r_q$-related edges from two graphs:
{\small$\triples_{\it r}^{r_q} \leftarrow \{(e_i, r, e_j) \in \triples_{\it r} \mid e_i = r_q \lor e_j = r_q\}$}
{\small$\triples_{\it p}^{r_q} \leftarrow \{(e_i, r, e_j) \in \triples_{\it p} \mid e_i = r_q \lor e_j = r_q\}$}

Expand potential edges:
{\small$\triples_{\it ex}^{r_q} \leftarrow \textit{SupportExpanding}(r_q, \triples_{\it r})$}

Identify unmatched edges in $\graph_{\it p}$:
{\small $\triples_{\it m} \leftarrow (\triples_{\it p}-\triples_{\it p}^{r_q})-(\triples_{\it r}-\triples_{\it r}^{r_q})$}

Filter prompting edges of $r_q$:
{\small $\triples_{\it conf}^{r_q} \leftarrow \textit{ConflictFiltering}(\triples_{\it ex}^{r_q} \cup \triples_{\it p}^{r_q}, \triples_{\it m},\beta)$}

Inject prompting edges into $\graph_{\it r}$: 
{\small $\hat{\graph}_{\it r}^{r_q} \leftarrow \{\rels, \mathcal{R}_{\it fund}, \triples_{\it conf}^{r_q} \cup \triples_{\it r}\}$  }
\end{algorithm}

\textbf{Type Conflict Filtering:}
In a specific KG, two relations sharing an entity type in semantics may link to two disjoint entity sets, thereby the interaction edge between them is mistaken. 
Because $r_q$ lacks support triples, we can only detect whether the relations $\rels_s$, having the same entity type as $r_q$ at side $s$, have conflicts with each other.
Therefore, we compare the $\rels_s$-involved edges in $\graph_{\it r}$ and $\graph_{\it p}$, and extract unmatched edges $\triples_{m}$. 
If a relation has many unmatched edges with other relations in $\rels_s$, it is less likely connected with $r_q$.
The conflict determination of each relation is defined as follows:
\begin{align}
\small
C(r_j) = \left|\{r_j|(r_j, i, r_k) \in  \triples_{m}, r_j,r_k \in \ents_{\it p}\}\right| \geq \beta \nonumber
\end{align}
The threshold hyperparameter $\beta$ determines the maximum accepted number of unmatched edges for each relation $r_j$. If $C(r_j)$ is true, the interaction edges connecting $r_j$ and $r_q$ would be removed.
\section{Experiments}
\label{sec:4}
\vspace{-2mm}

\begin{table*}[h]
\scriptsize
\centering
\setlength{\tabcolsep}{0.5em}
\begin{tabular}{l|cccc|cccc|cccc|c}
\hline
\textbf{Model} & \multicolumn{1}{l}{\textbf{FB:v1}} & \multicolumn{1}{l}{\textbf{FB:v2}} & \multicolumn{1}{l}{\textbf{FB:v3}} & \multicolumn{1}{l|}{\textbf{FB:v4}} & \multicolumn{1}{l}{\textbf{WK:v1}} & \multicolumn{1}{l}{\textbf{WK:v2}} & \multicolumn{1}{l}{\textbf{WK:v3}} & \multicolumn{1}{l|}{\textbf{WK:v4}} & \multicolumn{1}{l}{\textbf{NL:v1}} & \multicolumn{1}{l}{\textbf{NL:v2}} & \multicolumn{1}{l}{\textbf{NL:v3}} & \multicolumn{1}{l|}{\textbf{NL:v4}} & \multicolumn{1}{l}{\textbf{AVG}} \\
\hline
NBFNet & 9.2 & 1.9 & 1.3 & 0.0 & 3.3 & 0.1 & 0.0 & 0.0 & 0.0 & 1.4 & 0.5 & 0.0 & 1.5\\
REDGNN & 6.8 & 6.6 & 11.0 & 6.5 & 0.0 & 0.3 & 0.0 & 0.0 & 8.1 & 7.1 & 3.6 & 2.5 & 4.4 \\
DEqInGram & 7.6 & 0.9 & 7.0 & 1.1 & 0.6 & 0.2 & 0.5 & 5.2 & 6.8 & 4.6 & 4.3 & 7.4 & 3.9 \\
ISDEA & 1.8 & 1.3 & 0.9 & 0.8 & 1.4 & 0.3 & 1.2 & 0.3 & 3.5 & 3.1 & 3.5 & 1.9 & 1.7 \\
InGram & 9.8 & 6.2 & 11.4 & 7.0 & 7.0 & 0.2 & 2.2 & 2.3 & 12.2 & 12.4 & 13.8 & 14.9 & 8.3\\
ULTRA(3g) & 35.9 & 33.4 & 33.5 & 31.9 & 43.0 & 9.2 & 17.2 & \textbf{11.4} & 27.5 & 29.1 & 32.4 & 36.9 & 28.5 \\
ULTRA(4g) & 40.4 & 35.8 & 32.7 & 31.6 & 13.8 & 6.3 & 17.6 & 9.2 & 25.8 & 25.5 & 26.5 & 30.0 & 24.6 \\
ULTRA(50g) & 36.8 & 35.4 & 32.9 & 31.0 & 15.7 & 5.5 & 16.6 & 8.6 & 22.7 & 22.0 & 21.9 & 26.2 & 22.9 \\
\hline
Our(Llama2-7B) & 51.3 & \textbf{48.0} & \textbf{40.8} & 37.7 & 46.1 & 10.8 & 18.6 & 11.1 & 33.5 & \textbf{37.2} & 33.5 & 39.4 & 34.0 \\
Our(Llama2-13B) & 49.2 & 46.8 & 40.5 & 35.8 & \textbf{46.3} & 11.6 & 18.6 & 11.0 & 32.0 & 36.7 & 35.4 & 40.3 & 33.7 \\
Our(Mistral-7B) & 49.4 & 46.7 & 40.2 & 36.7 & 46.1 & 11.5 & \textbf{18.8} & 10.8 & 32.3 & 33.3 & 35.4 & 39.4 & 33.4 \\
Our(GPT-3.5) & 50.5 & 47.1 & \textbf{40.8} & \textbf{38.0} & 45.5 & 11.5 & 18.7 & 11.3 & \textbf{34.2} & 35.1 & \textbf{35.5} & \textbf{41.1} & \textbf{34.1} \\
Our(GPT-4) & \textbf{51.7} & 47.3 & 40.4 & 37.5 & 45.6 & \textbf{11.8} & \textbf{18.8} & 11.2 & 33.0 & 35.2 & 33.8 & 39.8 & 33.8 \\
\hline
\end{tabular}
\vspace{-3mm}
\caption{3-shot inductive reasoning results on three series of datasets, evaluated with Hits@10 (\%).}
\vspace{-1mm}
\label{table:3shot}
\end{table*}

\begin{table*}[h]
\scriptsize
\centering
\setlength{\tabcolsep}{0.5em}
\begin{tabular}{l|cccc|cccc|cccc|c}
\hline
\textbf{Model} & \multicolumn{1}{l}{\textbf{FB:v1}} & \multicolumn{1}{l}{\textbf{FB:v2}} & \multicolumn{1}{l}{\textbf{FB:v3}} & \multicolumn{1}{l|}{\textbf{FB:v4}} & \multicolumn{1}{l}{\textbf{WK:v1}} & \multicolumn{1}{l}{\textbf{WK:v2}} & \multicolumn{1}{l}{\textbf{WK:v3}} & \multicolumn{1}{l|}{\textbf{WK:v4}} & \multicolumn{1}{l}{\textbf{NL:v1}} & \multicolumn{1}{l}{\textbf{NL:v2}} & \multicolumn{1}{l}{\textbf{NL:v3}} & \multicolumn{1}{l|}{\textbf{NL:v4}} & \multicolumn{1}{l}{\textbf{AVG}} \\
\hline
NBFNet & 8.7 & 1.8 & 1.3 & 0.0 & 1.9 & 0.1 & 0.0 & 0.0 & 0.0 & 7.1 & 0.4 & 0.0 & 1.8\\
REDGNN & 6.3 & 6.4 & 10.3 & 5.4 & 0.2 & 0.1 & 0.4 & 0.0 & 6.3 & 6.4 & 1.9 & 0.9 & 3.7 \\
DEqInGram & 6.3 & 0.7 & 5.0 & 1.3 & 0.4 & 0.2 & 0.4 & 1.2 & 9.6 & 4.3 & 3.5 & 6.6 & 3.3 \\
ISDEA & 2.2 & 0.8 & 0.7 & 0.6 & 1.1 & 0.2 & 1.1 & 0.1 & 2.6 & 2.7 & 2.3 & 1.8 & 1.4 \\
InGram & 6.9 & 5.7 & 6.2 & 4.9 & 8.0 & 0.3 & 1.4 & 0.9 & 9.6 & 8.2 & 8.5 & 11.6 & 6.0 \\
ULTRA(3g) & 27.0 & 25.6 & 23.5 & 20.4 & 24.9 & 4.2 & 10.1 & 2.9 & 21.6 & 19.4 & 22.3 & 25.5 & 19.0 \\
ULTRA(4g) & 29.6 & 27.3 & 23.0 & 22.1 & 4.3 & 2.5 & 10.1 & 2.0 & 18.3 & 17.1 & 15.1 & 16.0 & 15.6 \\
ULTRA(50g) & 27.5 & 25.4 & 23.0 & 23.5 & 6.5 & 2.3 & 9.6 & 2.0 & 18.5 & 17.6 & 15.5 & 18.6 & 15.8 \\
\hline
Our(Llama2-7B) & 44.2 & \textbf{41.7} & 36.3 & 29.8 & 37.1 & 5.6 & 11.6 & 2.9 & 29.5 & 27.7 & 24.7 & 31.1 & 26.9 \\
Our(Llama2-13B) & 44.5 & 40.4 & 35.0 & 28.0 & \textbf{38.4} & 5.4 & \textbf{11.9} & 3.4 & 27.2 & 27.9 & 25.6 & 31.2 & 26.6 \\
Our(Mistral-7B) & 44.1 & 40.4 & 35.9 & 29.5 & 38.1 & 6.1 & \textbf{11.9} & 3.1 & 27.0 & 25.9 & \textbf{26.8} & 28.5 & 26.4 \\
Our(GPT-3.5) & 43.8 & \textbf{41.7} & 36.9 & \textbf{32.4} & 38.1 & 5.8 & 11.8 & \textbf{3.5} & 29.2 & \textbf{29.1} & 26.6 & \textbf{32.2} & \textbf{27.6} \\
Our(GPT-4) & \textbf{45.1} & 40.9 & \textbf{37.0} & 31.9 & 37.7 & \textbf{6.3} & 11.8 & \textbf{3.5} & \textbf{29.7} & 26.7 & 26.0 & 30.0 & 27.2 \\
\hline
\end{tabular}
\vspace{-3mm}
\caption{1-shot inductive reasoning results on three series of datasets, evaluated with Hits@10 (\%).}
\vspace{-5mm}
\label{table:1shot}
\end{table*}

We extensively evaluate our method on $K$-shot inductive reasoning of KGs. In particular, we wish to answer the following research questions: 
\textbf{RQ1}: How effective is our method under the low-resource scenarios of distinct KGs?
\textbf{RQ2}: \kaiupdate{How do different LLM prompts impact prompt graphs and KG inductive reasoning?} 
\textbf{RQ3}: How do the main components of our method impact the performance? 
\textbf{RQ4}: \kaiupdate{How does our method perform in more shots and full-shot scenarios?}
\textbf{RQ5}: How efficient is our model compared with traditional approaches? 
\textbf{RQ6}: How does GNN reasoning change before and after injecting the prompt graph in case studies?
\kaiupdate{Due to the space limitation, discussions about RQ5 and RQ6 are detailed in the Appendix.}

\vspace{-2mm}
\subsection{Experimental Setup}

\textbf{Low-resource Datasets.} 
We conduct $K$-shot inductive reasoning experiments on 108 low-resource datasets, constructed upon the 12 datasets used by the InGram work~\cite{ingram}.
The InGram datasets were derived from three real-world KG benchmarks: FB15k237 \cite{FB15k237}, Wikidata68K~\cite{TextKGR-RAILD-IJCKG22}, and NELL-995~\cite{NELL995}, abbreviated as FB, WK, and NL. 
There are four KG datasets (we call v1-v4) for each benchmark. 
For FB and NL datasets, we utilize the word-segmented relation names as the short description, while extracting official relation descriptions from WikiData for WK datasets.
Both structural and textual statistics for these datasets can be found in Table \ref{table:ds-test} and Table \ref{table:ds-text} in the Appendix.
Based on the InGram datasets, we create $K$-shot datasets tailored for 3-shot, 1-shot, and 0-shot scenarios. 
In each InGram dataset, we randomly retain $K$ support triples for each query relation $r_q$ and mask others when reasoning from $\ginf^{r_q}$. 
Ensuring a robust evaluation of our model, we perform this sampling process three times to create three variants for each few-shot dataset. 

\textbf{Baselines and Implementation.} 
We compare \method with eight baseline methods.
NBFNet \cite{NBF-NIPS21}, RED-GNN \cite{REDGNN-WWW22}, InGram \cite{ingram}, DEqInGram \cite{isdea}, and ISDEA \cite{isdea} are state-of-the-art GNN-based inductive models. 
The recent pre-trained model ULTRA($X$g) \cite{ultra} has three variants (3g, 4g, 50g), of which $X$ denotes the amount of pre-trained KG datasets.
Previous few-shot link prediction and text-based methods are neglected because they cannot perform on our task settings.
For \method, we train the GNN reasoner following the settings of ULTRA(3g) and employ five LLMs in the LLM prompter, including Llama2-7B, Llama2-13B \cite{Llama2}, 
Mistral-7B \cite{MINERVA-NIPS17}, GPT-3.5 \cite{Gpt3}, and GPT-4 \cite{Gpt4}. 
We utilize two evaluation metrics, MRR (Mean Reciprocal Rank) and Hits@N. 
Hyperparameters are selected via grid search according to the metrics on the validation set.
All experiments are performed on Intel Xeon Gold 6238R CPU @ 2.20GHz and $4\times$ NVIDIA RTX A30 GPUs.
Implementation details and hyperparameter configurations are shown in Appendix \ref{app:md}.

\begin{table*}[h]
\scriptsize
\centering
\setlength{\tabcolsep}{0.5em}
\begin{tabular}{l|cccc|cccc|cccc|c}
\hline
\textbf{Model} & \multicolumn{1}{l}{\textbf{FB:v1}} & \multicolumn{1}{l}{\textbf{FB:v2}} & \multicolumn{1}{l}{\textbf{FB:v3}} & \multicolumn{1}{l|}{\textbf{FB:v4}} & \multicolumn{1}{l}{\textbf{WK:v1}} & \multicolumn{1}{l}{\textbf{WK:v2}} & \multicolumn{1}{l}{\textbf{WK:v3}} & \multicolumn{1}{l|}{\textbf{WK:v4}} & \multicolumn{1}{l}{\textbf{NL:v1}} & \multicolumn{1}{l}{\textbf{NL:v2}} & \multicolumn{1}{l}{\textbf{NL:v3}} & \multicolumn{1}{l|}{\textbf{NL:v4}} & \multicolumn{1}{l}{\textbf{AVG}} \\
\hline
NBFNet & 8.7 & 1.8 & 1.3 & 0.0 & 1.9 & 0.1 & 0.0 & 0.0 & 0.0 & 7.1 & 0.4 & 0.0 & 1.8\\
REDGNN & 6.3 & 6.4 & 10.3 & 5.4 & 0.2 & 0.1 & 0.0 & 0.0 & 6.0 & 6.4 & 0.8 & 2.9 & 3.7 \\
DEqInGram & 0.3 & 0.4 & 1.3 & 0.7 & 2.3 & 0.1 & 0.0 & 0.1 & 4.0 & 3.2 & 2.2 & 2.0 & 1.4 \\
ISDEA & 1.7 & 0.1 & 0.2 & 0.2 & 0.2 & 0.0 & 0.4 & 0.0 & 0.9 & 1.4 & 0.3 & 1.2 & 0.6 \\
InGram & 1.7 & 4.3 & 5.8 & 2.7 & 5.4 & 0.1 & 0.2 & 0.2 & 7.0 & 7.0 & 2.9 & 8.1 & 3.8 \\
ULTRA(3g) & 14.5 & 12.8 & 10.1 & 9.8 & 1.9 & 0.8 & 1.3 & 0.4 & 11.3 & 6.8 & 6.1 & 6.0 & 6.8 \\
ULTRA(4g) & 15.5 & 14.8 & 10.1 & 10.7 & 1.8 & 1.0 & 1.3 & 0.4 & 9.6 & 7.1 & 5.9 & 6.1 & 7.0 \\
ULTRA(50g) & 10.3 & 10.0 & 6.3 & 10.0 & 2.3 & 1.2 & 1.3 & 0.4 & 8.9 & 7.7 & 5.8 & 4.6 & 5.7 \\
\hline
Our(Llama2-7B) & 19.3 & 20.3 & 13.3 & 15.3 & 2.9 & 1.0 & 1.6 & 0.7 & 18.0 & \textbf{19.0} & 9.5 & 16.1 & 11.4 \\
Our(Llama2-13B) & 20.9 & 17.9 & 13.6 & 12.1 & 7.8 & 1.1 & 1.6 & 1.6 & \textbf{19.1} & 18.4 & \textbf{15.5} & 14.8 & 12.0 \\
Our(Mistral-7B) & 25.4 & 25.4 & 22.2 & 19.5 & 4.3 & 1.2 & 1.7 & 1.1 & 17.2 & 15.6 & 14.3 & 11.9 & 13.3 \\
Our(GPT-3.5) & \textbf{30.0} & \textbf{26.7} & 24.0 & \textbf{22.5} & 30.4 & 2.0 & 3.0 & 1.4 & 17.8 & 17.8 & 15.3 & 17.7 & \textbf{17.4} \\
Our(GPT-4) & 28.4 & 25.1 & \textbf{25.6} & 21.3 & \textbf{35.0} & \textbf{2.4} & \textbf{3.1} & \textbf{1.8} & 18.3 & 14.0 & 14.7 & \textbf{19.1} & \textbf{17.4} \\
\hline
\end{tabular}
\vspace{-2mm}
\caption{0-shot inductive reasoning results on three series of datasets, evaluated with Hits@10 (\%).}
\label{table:0shot}
\vspace{-3mm}
\end{table*}

\subsection{Main Experimental Results (RQ1)}
We compare \method with baselines on inductive reasoning tasks under the 3-shot, 1-shot, and 0-shot settings.
We report the average Hits@10 results over three variants of each $K$-shot InGram dataset in Table \ref{table:3shot}, Table \ref{table:1shot}, and Table \ref{table:0shot}. 
More detailed results on other metrics can be found in the Appendix. 
We observe that the first four GNN-based methods underperform pre-trained ULTRA and our method.
NBFNet and REDGNN struggle with unseen relations, while InGram, ISDEA, and DEqInGram rely on sufficient support triples to form a similar distribution of node degrees. 
Conversely, ULTRA excels in low-resource settings due to its pre-training on multiple KGs, especially in the 3-shot scenario, though more KG pre-training in ULTRA(50g) doesn't markedly boost performance.
Pre-trained on three KGs like ULTRA(3g), \method significantly outperforms previous methods by a wide margin on average. 
Notably, in the 0-shot setting, the average Hits@10 for Our(GPT-4) method is twice as high as that of ULTRA, highlighting the effectiveness of our prompting paradigm.
In the comparison of different LLMs, GPT-3.5 and GPT-4 exhibit superior performance in the 0-shot setting. 
Due to our simplification for LLM requests, the lightweight Llama2-7B already performs well in few-shot scenarios, which indicates the robustness of our method.

\subsection{Performance of Different Prompts (RQ2)}
\begin{figure}[!th]
\centering
\vspace{-2mm}
\includegraphics[width=0.47\textwidth]{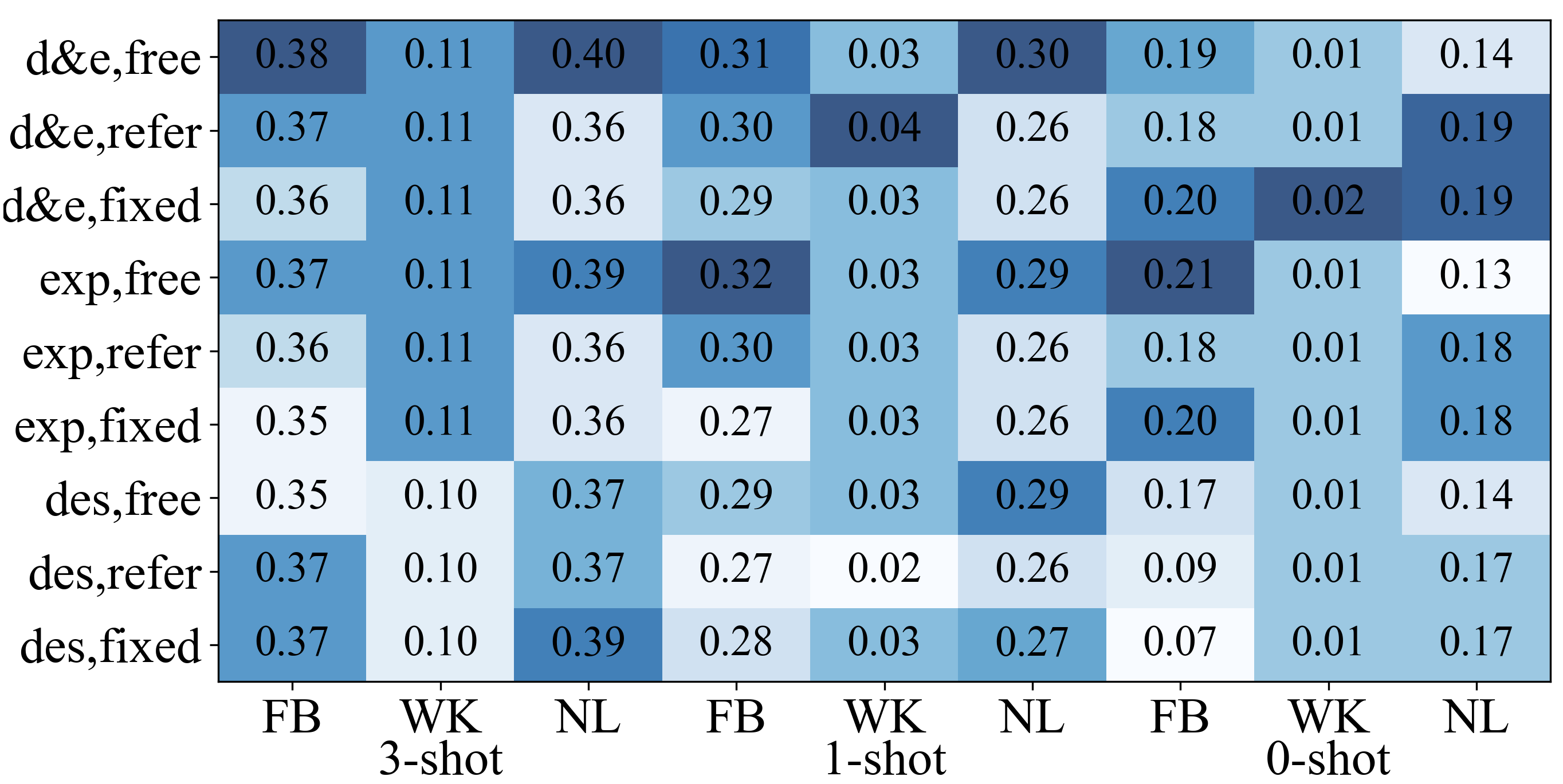}
\vspace{-2mm}
\caption{Hits@10 results with different prompt settings in GPT-4. Darker colors indicate higher values.}
\label{fig:prompt1}
\vspace{-4mm}
\end{figure}

To generate the prompt graph, we employ a series of textual prompts to guide the LLM prompter.  
As shown in Figure \ref{fig:prompt1}, we compare the performance of Our(GPT-4) on the three $v4$ datasets by varying the \textit{Relation Information Form} and \textit{Output Entity Type}, respectively.
The `d\&e' form outperforms the other two in most scenarios, because textual descriptions reflect the core features while two entity names indicate the direction of the relation. Regarding entity types, the `free' setting works better in few-shot scenarios. In zero-shot datasets, prompts using (`d\&e', `fixed') outperform the others.

In Figure \ref{fig:prompt2} in Appendix, we analyse the effectiveness of the prompt graph by calculating the F1 metric between the prompt interaction edges and ground-truth edges in the raw KGs.
The trend of F1 metrics across different prompts is similar to that of Hits@10, it shows that a prompt graph closer to ground truth leads to better performance.
The results indicate the importance of prompt settings and more refined prompts would be our future work.

\begin{table}[]
\small
\centering
\scriptsize
\begin{tabular}{llccc}
\hline
 & \textbf{Model} & \textbf{FB:v4} & \textbf{WK:v4} & \textbf{NL:v4} \\
 \hline
\multirow{6}{*}{\textbf{3-shot}} & Our(GPT-4) & 37.5 & \textbf{11.2} & \textbf{38.8} \\
 & with ULTRA & 34.6 & 11.0 & 37.1 \\
 & with OODKG & 36.8	& 10.6	& 35.9 \\
 & w/o Prompt & \textbf{37.7} & 11.1 & 37.7 \\
 & w/o Calibrator & 37.5 & 10.6 & 37.4 \\
 & w/o LLM & 30.9 & 10.1 & 33.4 \\
 \hline
\multirow{6}{*}{\textbf{1-shot}} & Our(GPT-4) & 31.9 & \textbf{3.5} & \textbf{29.3} \\
 & with ULTRA & 28.1 & \textbf{3.5} & 27.0 \\
 & with OODKG & \textbf{32.3}	& 3.2 & 26.6 \\
 & w/o Prompt & 28.0 & 2.6 & 25.9 \\
 & w/o Calibrator & 31.1 & 2.8 & \textbf{29.3} \\
 & w/o LLM & 23.8 & 2.6 & 25.7 \\
 \hline
\multirow{6}{*}{\textbf{0-shot}} & Our(GPT-4) & \textbf{21.3} & \textbf{1.8} & \textbf{18.5} \\
 & with ULTRA & 18.2 & 1.4 & 15.5 \\
 & with OODKG & 21.2	& 1.3 & 18.3 \\
 & w/o Prompt & 8.9 & 0.4 & 3.1 \\
 & w/o Calibrator & 20.2 & 1.4 & \textbf{18.5} \\
 & w/o LLM & 8.9 & 0.4 & 3.1 \\
\hline
\multirow{2}{*}{\textbf{2-shot}} & Our(GPT-4) & \textbf{35.6} & \textbf{5.4} & \textbf{37.1} \\
 & ULTRA(3g) & 28.7 & 5.1 & 31.8 \\
 \hline
\multirow{2}{*}{\textbf{4-shot}} & Our(GPT-4) & \textbf{39.6} & 7.3 & \textbf{45.4} \\
 & ULTRA(3g) & 36.7 & \textbf{7.7} & 42.1 \\
 \hline
\multirow{2}{*}{\textbf{8-shot}} & Our(GPT-4) & \textbf{45.0} & 9.8 & \textbf{47.2} \\
 & ULTRA(3g) & 42.8 & \textbf{10.1} & 45.2 \\
 \hline
\multirow{2}{*}{\textbf{16-shot}} & Our(GPT-4) & \textbf{46.6} & \textbf{17.4} & \textbf{55.2} \\
 & ULTRA(3g) & 46.2 & 17.2 & 53.8 \\
\hline
\multirow{5}{*}{\textbf{full-shot}} & Our(GPT-4) & 63.1 & 28.5 & \textbf{67.7} \\
& with Llama2-7B & \textbf{63.3} & 27.8 & \textbf{67.7} \\
& w/o Prompt & 63.1 & 26.9 & 67.1 \\
& InGram & 37.1 & 16.9 & 50.6 \\
& ULTRA(3g) & 62.9 & \textbf{28.9} & 63.2 \\
 \hline
\end{tabular}
\vspace{-2mm}
\caption{Ablation studies evaluated with Hits@10 (\%).}
\vspace{-5mm}
\label{table:abla}
\end{table}

\vspace{-2mm}
\subsection{Ablation Studies (RQ3, RQ4)}
To validate the impact of the three components on model performance, we conduct \textbf{ablation experiments} on \method with GPT-4, as shown in Table \ref{table:abla}.
The variant `with ULTRA' employs ULTRA(3g) to replace our pre-trained GNN reasoner and the performance decline indicates the effect of our enhancing techniques.
`w/o Prompt' denotes utilizing the pre-trained KG reasoner directly without prompt graphs, which performs well in 3-shot scenarios but struggles with fewer-shot ones.  
The next two variants prove that the LLM prompter and the prompt calibrator both enhance the model’s performance.
\kaiupdate{\textbf{`with OODKG'} denotes pre-training the GNN reasoner with three KGs out of the evaluation KG distribution, including YAGO3-10 \cite{yago3}, DBPedia100k \cite{dbp100k}, and WN18RR \cite{WN18RR} Its competitive performance indicates the robustness of our pre-training strategy.}

\kaiupdate{
Regarding RQ4, \method consistently surpasses ULTRA(3g) in most cases of 2/4/8/16 shots experiments. As the full results presented in Table \ref{table:moreshot} in Appendix, the average performance improvement is still significant in more-shot scenarios, which validates our model's superiority.
We also verify \method in the \textbf{full-shot scenarios}. Our(GPT-4) and Our(Llama2-7b) outperform ULTRA(3g) and InGram on FB:v4 and NL:v4 datasets. The results prove the good applicability of our method in practice.}

\section{Conclusions}
\label{sec:6}

We propose a novel pre-training and prompting framework, \method, for low-resource inductive reasoning across arbitrary KGs. 
To generate an effective prompt graph for few-shot relation types, we design enhancing techniques for the GNN reasoner and instruction prompts for the LLM prompter. Besides, a novel prompt calibrator is proposed to mitigate the potential noise and achieve the information alignment of the above two components. 
Extensive experiments have verified that the \method achieves significant performance in both few-shot and zero-shot scenarios. 
Besides, \method requires no model fine-tuning, thereby having advantages of better efficiency and scalability than previous GNN-based methods.

\clearpage
\section{Limitations.}
Here, we discuss two potential limitations of \method. 
First, unlike recent GNN-based fully inductive methods, our method requires a brief relational context for each relation type. Despite our efforts to minimize text requirements, it may be unavailable in certain scenarios without manual inputs from users.
Second, the usage of textual context in our method is not sufficient enough. To circumvent the need for additional model training, we streamline the LLM queries by only asking for entity types. 
Incorporating semantic embeddings into GNNs or fine-tuning the LLM prompter to leverage relation semantics could further enhance performance. 
Exploring this avenue will constitute a primary focus of our future work.
The potential risks of our work may include generating factual triples about privacy or fake information, depending on the legality and reliability of the input data.
In addition, we utilize the ChatGPT AI assistant when polishing some paragraphs of the paper draft.

\section*{Acknowledgments}
This research is supported by the Ministry of Education, Singapore, under its AcRF Tier-2 Grant (T2EP20122-0003). Any opinions, findings and conclusions or recommendations expressed in this material are those of the author(s) and do not reflect the views of the Ministry of Education, Singapore.

\vspace{-4mm}
\bibliography{bibtex_url,bibtex_url2}

\clearpage
\appendix

\section{Implementation Details and Hyperparameters}
\label{app:md}
\vspace{-5mm}

We introduce the statistics of pre-training and evaluation datasets in Table \ref{table:ds-tra}, Table \ref{table:ds-test}.
Following previous work \cite{REDGNN-WWW22, NBF-NIPS21}, we augment the triples in each $\graph$ with reverse and identity relations. The augmented triple set $\triples^+$ is defined as: $\triples^+ = \triples \cup \{(e_t, r^{-1}, e_h)|(e_h, r, e_t) \in \triples\} \cup \{(e, r^i, e)|e\in  \ents\}$, where the relation $r^{-1}$ is the reverse relation of a relation $r$, the relation $r^i$ refers to the identity relation, and the number of augmented triples is $|\triples^+| = 2|\triples|+|\ents|$.

\textbf{Baseline Implementation Details.} 
We train NBFNet, RED-GNN, and ISDEA using the training graph of InGram datasets \cite{ingram} and evaluate the low-resource datasets. When evaluating InGram, DEqInGram, and ULTRA, we utilize their officially provided checkpoints directly.
Although there are some previous baselines \cite{NodePiece,DRUM-NIPS19,GraIL-ICML19}, they exhibit near-zero performance in original full-shot inductive tasks reported by InGram \cite{ingram}. Therefore, we ignore them in this more challenging low-resource reasoning task.

\textbf{\method Implementation Details.} 
We train our KG reasoner with three commonly-used KG datasets, WN18RR \cite{WN18RR}, FB15k237 \cite{FB15k237}, and CodexM~\cite{Codex}, following the settings of ULTRA(3g). 
Concerns about potential relation leakage during pre-training can be ignored because neither our method nor ULTRA learns relation-specific parameters. Moreover, in low-resource settings, the available relational information is limited, and markedly distinct from the original data.
We employ five popular LLMs as the LLM prompter, including Llama2-7B, Llama2-13B \cite{Llama2}, 
Mistral-7B \cite{MINERVA-NIPS17}, GPT-3.5 \cite{Gpt3}, and GPT-4 \cite{Gpt4}.  
Llama2 and Mistral-7B are hosted directly on our servers, while the GPT-3.5 and GPT-4 models are accessed through the OpenAI API.
For Prompt Calibrater, we set the loss balancing ratio $\alpha$ to 0.5, and the low-resource masking ratio $\gamma$ to 0.1. The filtering threshold $\beta$ is selected from $\{1,3,5,\text{Mean},\text{Max}\}$, in which the latter two are calculated from values of all relation conflicts.

All experiments are performed on Intel Xeon Gold 6238R CPU @ 2.20GHz and NVIDIA RTX A30 GPUs (four for pretraining and one for evaluation), and are implemented in Python using the PyTorch framework.
Our source code is implemented based on ULTRA\footnote{https://github.com/DeepGraphLearning/ULTRA}, which is available under the MIT License. 
We utilize the Llama2 and Mistral models under the corresponding licenses and call the GPT-3.5 and GPT-4 APIs obeying OpenAI terms.
All employed KG datasets are open and commonly used.

\section{Detailed Discussion about Prompt Calibrator}
\label{app:ddpc}
Here, we further introduce the algorithm in prompt calibrator, accompanied by specific examples. This algorithm aims to extract high-confidence prompting edges while minimizing potential noise edges by harnessing ground-truth information from the existing topological relation graph.

The mechanism for few-shot support expanding focuses on the ground-truth interaction edges of the target few-shot relation $r_q$. This process follows two rules, enabling the inference of more potential edges from these ground-truth edges.
The first rule extends the entity type set of $r_q$ by incorporating the entity types of its linked relations. For instance, in Figure 1, if we observe that the `co-worked' relation has a `h2t' edge to `sung by', the entity types on the tail side of `sung by' become more likely to be the head entity type of `co-worked'. These extended entity types facilitate the deduction of more potential edges according to the rules in prompt graph construction.
The second rule employs a specific 2-hop pattern. For instance, because `sung by' and `lyricized by' are linked by a `t2t' edge, they are likely to share at least one same entity type on the tail side. Consequently, we can infer that `co-worked' may have a `h2t' edge to `lyricized by'. However, these expanding edges remain uncertain and necessitate further noise filtering.

The type conflict filtering mechanism centers on the ground-truth interaction edges of other relations except $r_q$. As all entity types for each relation are predicted by the LLM prompter, prompting edges among these full-shot relations are also accessible. Since full-shot relations are presumed to encompass all interaction edges, disparities between prompting edges and ground-truth edges can indicate the accuracy of edge type prediction.
Hence, given a set of relations sharing the same entity type as $r_q$, we can analyze the mistaken prompting edges among them and assess whether the entity range of each relation is disjoint from that of others. Such relations with conflicts are more likely to have no connections with $r_q$.
For example, `directed by' and `lyricized by' share the same tail entity type 'people', but they do not share the same tail entities. Consequently, these two relations are hardly to be simultaneously linked to `sung by'.

\begin{table}[!ht]
\scriptsize
\centering
\vspace{-2mm}
\begin{tabular}{lccccc}
\hline
\multirow{2}{*}{Dataset} & \multirow{2}{*}{$|\etrain|$} & \multirow{2}{*}{$|\rtrain|$} & \multicolumn{3}{c}{$|\ttrain|$} \\
& & & {\#Train} & {\#Validation} & {\#Test} \\
\hline
WN18RR & 40.9k & 11 & 86.8k & 3.0k & 3.1k \\
FB15k-237 & 14.5k & 237 & 272.1k & 17.5k & 20.4k \\
CodexMedium & 17.0k & 51 & 185.5k & 10.3k & 10.3k \\
\hline
\end{tabular}
\vspace{-5px}
\caption{Statistics of pre-training KG datasets.}
\label{table:ds-tra}
\vspace{-4mm}
\end{table}
\begin{table}[hb]
\scriptsize
\centering
\setlength{\tabcolsep}{0.35em}
\vspace{-2mm}
\begin{tabular}{lccccccccc}
\hline
\multirow{2}{*}{} & \multicolumn{3}{c}{FB} & \multicolumn{3}{c}{WK} & \multicolumn{3}{c}{NL} \\
 & $|\einf|$ & $|\rinf|$ & $|\tinf|$ & $|\einf|$ & $|\rinf|$ & $|\tinf|$ & $|\einf|$ & $|\rinf|$ & $|\tinf|$ \\
\hline
v1 & 2146 & 120 & 3717 & 3228 & 74 & 5652 & 4097 & 216 & 28579 \\ 
v2 & 2335 & 119 & 4294 & 9328 & 93 & 16121 & 4445 & 205 & 19394 \\ 
v3 & 1578 & 116 & 3031 & 2722 & 65 & 5717 & 2792 & 186 & 15528 \\ 
v4 & 1709 & 53 & 3964 & 12136 & 37 & 22479 & 2624 & 77 & 11645 \\ 
\hline
\end{tabular}
\vspace{-5px}
\caption{Statistics of evaluating InGram datasets.}
\label{table:ds-test}
\vspace{-4mm}
\end{table}
\begin{table}[ht]
\scriptsize
\centering
\vspace{-2mm}
\begin{tabular}{clp{4cm}}
\hline
Dataset & Textual Form & Details \\
\hline
\multirow{6}{*}{FB} & \multirow{2}{*}{Description} & Average Words: 4.53 \\
& & Average Tokens: 61.56 \\
& (Example) & "/people/person/profession" \\
\cline{2-3}
& \multirow{2}{*}{Entity Name} &  Average Words: 2.30 \\
& & Average Tokens: 15.21 \\
& (Example) & "Stan Lee" \\
\hline
\multirow{7}{*}{WK} & \multirow{2}{*}{Description} & Average Words: 15.93 \\
& & Average Tokens: 103.42 \\
& (Example) & "residence: the place where the person is or has been, resident" \\
\cline{2-3}
& \multirow{2}{*}{Entity Name} &  Average Words: 2.57 \\
& & Average Tokens: 17.68 \\
& (Example) & "coquette (film)" \\
\hline
\multirow{6}{*}{NL} & \multirow{2}{*}{Description} & Average Words: 3.86 \\
& & Average Tokens: 24.46 \\
& (Example) & "person graduated from university" \\
\cline{2-3}
& \multirow{2}{*}{Entity Name} &  Average Words: 2.98 \\
& & Average Tokens: 20.91 \\
& (Example) & "city: portland" \\
\hline
\end{tabular}
\vspace{-5px}
\caption{Statistics of KG textual data.}
\label{table:ds-text}
\vspace{-4mm}
\end{table}


\begin{table}[ht]
\scriptsize
\centering
\begin{tabular}{cp{5cm}}
\hline
Prompt Setting & (FB, des, fixed) \\
\hline
Task Description & The dictionary rel\_dict includes brief information of partial relationships in a knowledge graph. Please analyze the possible entity types of each relationship's head and tail entities. \\
EntType & The candidate entity types are strictly fixed in ["genre/type", "person", "animal", "location/place", "organization", "creative work", "time", "profession", "event", "actual item", "language"]. \\
RelInfo & rel\_dict = \{"rel0": "music artist origin", "rel1": "film actor film. film performance film"\} \\
Output & 
Here are the results: 
\{
rel0: \{"head": ["person"], "tail": ["location/place"]\},
rel1: \{"head": ["person"], "tail": ["creative work"]\}
\} \\
\hline
Prompt Setting & (WK, exp, refer) \\
\hline
Task Description & The dictionary rel\_dict includes brief information of partial relationships in a knowledge graph. Please analyze the possible entity types of each relationship's head and tail entities. \\
EntType & The candidate entity types are not limited to ["genre/type", "person", "animal", "location/place", "organization", "creative work", "time", "profession", "event", "actual item", "language"]. \\
RelInfo & rel\_dict = \{"rel6": \{"head entity": "lesser asiatic yellow bat", "tail entity": "morphospecies"\}, "rel8": \{"head entity": "hexacinia", "tail entity": "peacock flies"\}\} \\
Output & 
Here are the results: 
\{
rel6: \{"head": ["animal"], "tail": ["morphospecies"]\},
rel8: \{"head": ["creative work"], "tail": ["genre/type"]\}
\} \\
\hline
Prompt Setting & (NL, d\&e, free) \\
\hline
Task Description & The dictionary rel\_dict includes brief information of partial relationships in a knowledge graph. Please analyze the possible entity types of each relationship's head and tail entities. \\
EntType & {\it None} \\
RelInfo & rel\_dict = \{ "rel0": \{"description": "sport fans in country", "head entity": "sport: skiing", "tail entity": "country: america"\}, "rel1": \{"description": "animal eat vegetable", "head entity": "bird: chickens", "tail entity": "vegetable: corn"\}\} \\
Output & 
Here are the results: 
\{
rel0: \{"head": ["sport"], "tail": ["country"]\},
rel1: \{"head": ["animal"], "tail": ["vegetable"]\}
\} \\
\hline
\end{tabular}
\vspace{-2mm}
\caption{Examples of textual prompts on three KGs.}
\label{table:proexp}
\vspace{-4mm}
\end{table}

\begin{table*}[t]
\small
\centering
\setlength{\tabcolsep}{0.2em}
\begin{tabular}{lc|cccc|cccc|cccc|c}
\hline
\textbf{Shot} & \textbf{Method} & \textbf{FB:v1} & \textbf{FB:v2} & \textbf{FB:v3} & \textbf{FB:v4} & \textbf{WK:v1} & \textbf{WK:v2} & \textbf{WK:v3} & \textbf{WK:v4} & \textbf{NL:v1} & \textbf{NL:v2} & \textbf{NL:v3} & \textbf{NL:v4} & \textbf{AVG} \\
\hline
16shot & ULTRA(3g) & 55.0 & 49.6 & 46.8 & 46.2 & 43.5 & 17.1 & 32.1 & 17.2 & 47.3 & 43.1 & 46.0 & 53.8 & 41.5 \\
16shot & Our(GPT-4) & \textbf{60.2} & \textbf{55.3} & \textbf{49.4} & \textbf{46.6} & \textbf{52.5} & \textbf{17.6} & \textbf{32.2} & \textbf{17.4} & \textbf{51.7} & \textbf{45.6} & \textbf{47.9} & \textbf{55.2} & \textbf{44.3} \\
8shot & ULTRA(3g) & 47.6 & 44.3 & 44.7 & 42.8 & 35.9 & 16.9 & 30.2 & \textbf{10.1} & 45.3 & 32.8 & \textbf{44.2} & 45.2 & 36.7 \\
8shot & Our(GPT-4) & \textbf{56.7} & \textbf{52.5} & \textbf{48.0} & \textbf{45.0} & \textbf{49.9} & \textbf{17.5} & \textbf{31.5} & 9.8 & \textbf{46.1} & \textbf{36.6} & 42.4 & \textbf{47.2} & \textbf{40.3} \\
4shot & ULTRA(3g) & 39.4 & 38.4 & 36.2 & 36.7 & 28.4 & 10.2 & 20.2 & \textbf{7.7} & 37.7 & 32.2 & 34.7 & 42.1 & 30.3 \\
4shot & Our(GPT-4) & \textbf{52.9} & \textbf{49.8} & \textbf{43.7} & \textbf{39.6} & \textbf{42.1} & \textbf{11.2} & \textbf{21.7} & 7.3 & \textbf{40.7} & \textbf{39.0} & \textbf{35.3} & \textbf{45.4} & \textbf{35.7} \\
2shot & ULTRA(3g) & 32.0 & 29.8 & 29.1 & 28.7 & 28.2 & 7.9 & 13.6 & 5.1 & 28.8 & 26.9 & 30.7 & 31.8 & 24.4 \\
2shot & Our(GPT-4) & \textbf{49.0} & \textbf{41.7} & \textbf{37.8} & \textbf{35.6} & \textbf{42.4} & \textbf{11.3} & \textbf{16.3} & \textbf{5.4} & \textbf{35.8} & \textbf{34.1} & \textbf{34.3} & \textbf{37.1} & \textbf{31.7} \\
\hline
\end{tabular}
\vspace{-2mm}
\caption{Experimental results with more shots, evaluated with Hits@10(\%).}
\vspace{-2mm}
\label{table:moreshot}
\end{table*}

\begin{table}[h]
\scriptsize
\centering
\begin{tabular}{llllll}
\hline
Dataset & K & LLM & Input & Output & $\beta$ \\
\hline
FB:v1 & 3 & gpt4 & des & refer & 1 \\
FB:v1 & 1 & gpt4 & des & refer & 1 \\
FB:v1 & 0 & gpt3.5 & exp & free & max \\
FB:v2 & 3 & gpt4 & des & refer & 1 \\
FB:v2 & 1 & llama7b & des & fixed & 5 \\
FB:v2 & 0 & gpt3.5 & exp & free & max \\
FB:v3 & 3 & llama7b & d\&e & fixed & 5 \\
FB:v3 & 1 & gpt4 & exp & free & 5 \\
FB:v3 & 0 & gpt4 & exp & fixed & max \\
FB:v4 & 3 & gpt3.5 & d\&e & free & max \\
FB:v4 & 1 & gpt3.5 & d\&e & free & max \\
FB:v4 & 0 & gpt3.5 & d\&e & refer & max \\
\hline
NL:v1 & 3 & gpt3.5 & des & free & max \\
NL:v1 & 1 & gpt4 & d\&e & free & max \\
NL:v1 & 0 & llama13b & d\&e & free & max \\
NL:v2 & 3 & llama7b & des & free & 3 \\
NL:v2 & 1 & gpt3.5 & des & free & 5 \\
NL:v2 & 0 & llama7b & d\&e & free & max \\
NL:v3 & 3 & gpt3.5 & des & free & max \\
NL:v3 & 1 & mistral7b & des & refer & max \\
NL:v3 & 0 & llama13b & exp & refer & max \\
NL:v4 & 3 & gpt3.5 & des & free & max \\
NL:v4 & 1 & gpt3.5 & des & free & max \\
NL:v4 & 0 & gpt4 & d\&e & refer & max \\
\hline
WK:v1 & 3 & llama13b & exp & free & max \\
WK:v1 & 1 & llama13b & exp & free & max \\
WK:v1 & 0 & gpt4 & d\&e & free & 3 \\
WK:v2 & 3 & gpt4 & exp & free & 1 \\
WK:v2 & 1 & gpt4 & d\&e & free & 3 \\
WK:v2 & 0 & gpt4 & d\&e & refer & mean \\
WK:v3 & 3 & gpt4 & d\&e & free & 3 \\
WK:v3 & 1 & mistral7b & d\&e & free & 1 \\
WK:v3 & 0 & gpt4 & d\&e & fixed & max \\
WK:v4 & 3 & gpt3.5 & des & free & max \\
WK:v4 & 1 & gpt4 & d\&e & refer & 1 \\
WK:v4 & 0 & gpt4 & d\&e & fixed & 5 \\
\hline
\end{tabular}
\vspace{-5px}
\caption{Hyperparameter settings of best Hits@10.}
\label{table:hyper}
\end{table}

\section{Efficiency Analysis (RQ5)}
The primary benefit of \method lies in its ability to infer new, arbitrary Knowledge Graphs (KGs) without the need for training. 
It stems from the fact that both the GNN reasoner and the LLM prompter remain parameter-frozen during the prompting process. 
As a result, the only computational overhead introduced is associated with the prompt graphs.
Except for the LLM requests for each relation in the KG, 
the prompt graph is generated only once for each few-shot relation type, whose computational complexity would not exceed $\mathcal{O}(|\rinf|^2)$. Given that the quantity of relations in most KGs is significantly lower than that of entities, the time and space costs in this process are negligible.
Specifically, the checkpoint file size of our KG reasoner is only 2.18 MB, which is similar to that of ULTRA (2.03 MB). The total pretraining time is around eight GPU hours for ten training epochs. The evaluation time on each dataset only costs several minutes.
Regarding the model scalability, \method can be applied in large-scale KGs the same as ULTRA \cite{ultra}. This work focuses on low-resource challenges, more experiments on large-scale datasets will be our future work.
\vspace{-2mm}
\section{Case Studies (RQ6)}
\vspace{-2mm}
We select several queries from the zero-shot NL:v4 dataset and compare the outputs of Our(GPT-4) with and without the prompt graph. 
The final relation graphs and top five outputted entities are shown in Figure \ref{fig:appcs}.
Generally, we observe that \method injects multiple prompt edges into the relation graph which directly changes the outputted entity ranking.
Most linked relations by prompt edges are reasonable, thereby improving the relative relation embedding for the query relation.

\begin{figure}[!th]
\centering
\vspace{-2mm}
\includegraphics[width=0.47\textwidth]{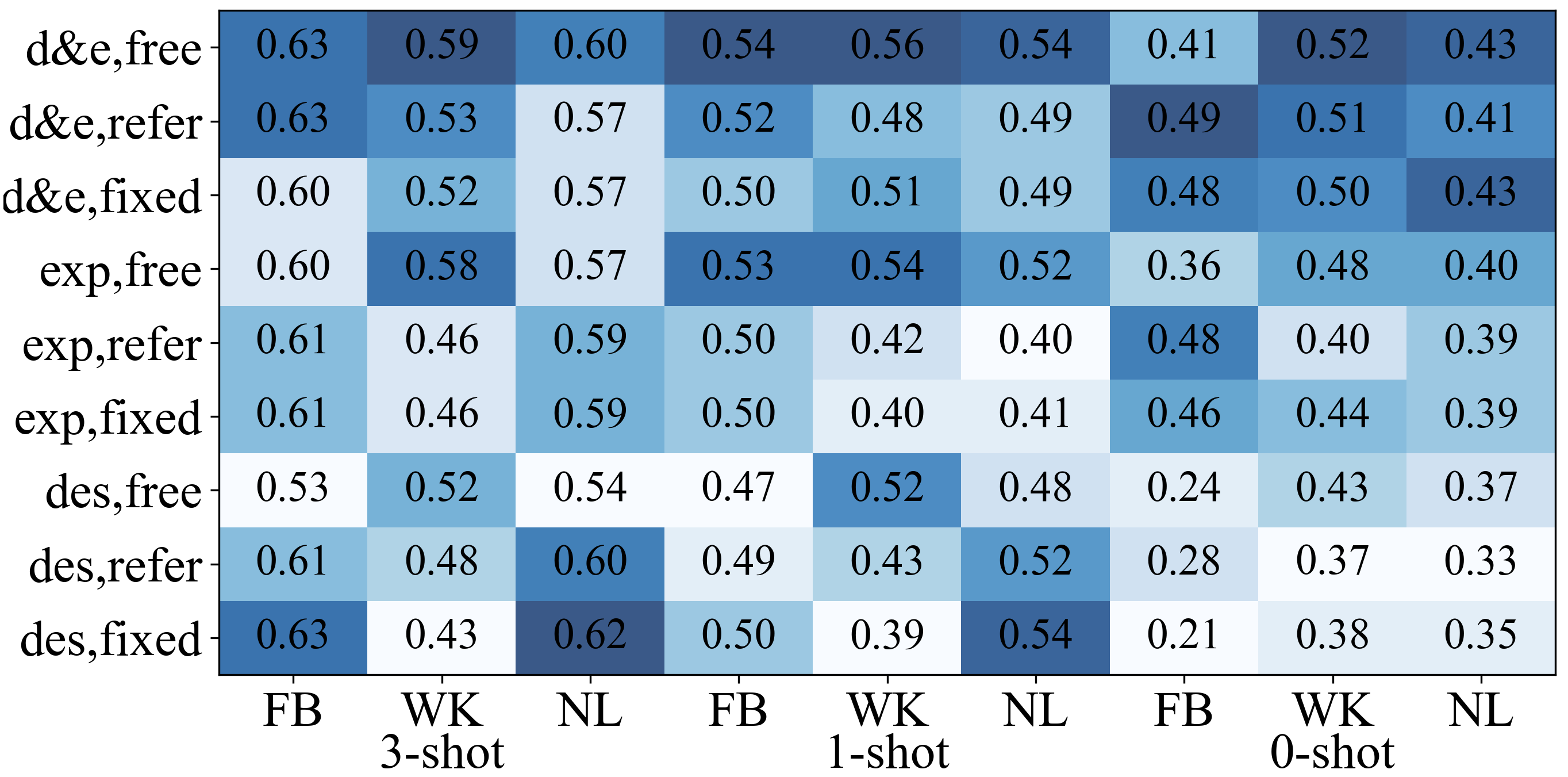}
\vspace{-2mm}
\caption{F1 metrics of interaction edges with different prompt settings in GPT-4. Darker colors indicate higher values.}
\label{fig:prompt2}
\vspace{-6mm}
\end{figure} 

\section{Detailed Related Work}
\label{sec:relatedwork}

\textbf{KG Inductive Reasoning:}
Traditional \emph{transductive} KG embedding models, represented by TransE \cite{TransE}, DistMult \cite{DistMult-paper} and RotatE \cite{RotatE}, learn continuous vectors in the embedding space to represent each entity and relation in the knowledge graph~\cite{2017Survey,OurComp,OurHaLE}. 
However, these models 
lack generalization capabilities for new entities within the same graph or across different KGs~\cite{OurMulDE,GoogleAttH}. 
In contrast, \emph{inductive} KG reasoning methods~\cite{NBF-NIPS21} overcome this limitation by generalizing to KGs with unseen entities or relations.
Most existing inductive methods \cite{Cycle-ICML22,InductivE-IJCNN21,IKR-JWWW23,TACT-AAAI21} leverage Graph Neural Networks (GNNs) to generate ``relative'' entity embeddings, by extracting local structural features from an induced graph of the query entity.
GraIL \cite{GraIL-ICML19} extracts an enclosing subgraph between the query entity and each candidate entity, but suffers from high computational complexity. 
NBFNet \cite{NBF-NIPS21} and RED-GNN \cite{REDGNN-WWW22} propagate query features through the $L$-hop neighborhood subgraph of the query entity. 
To further enhance model efficiency, recent studies have focused on algorithmic improvements, including path-pruning methods in the GNN propagation process \cite{AdaProp-Arxiv22,Astar-Arxiv23,OurGraPE}. 
These inductive methods struggle to generalize to KGs with new relation types, as the entity embeddings are still a function of a predetermined relation vocabulary.

\begin{figure*}[!t]
\centering
\includegraphics[width=0.9\textwidth]{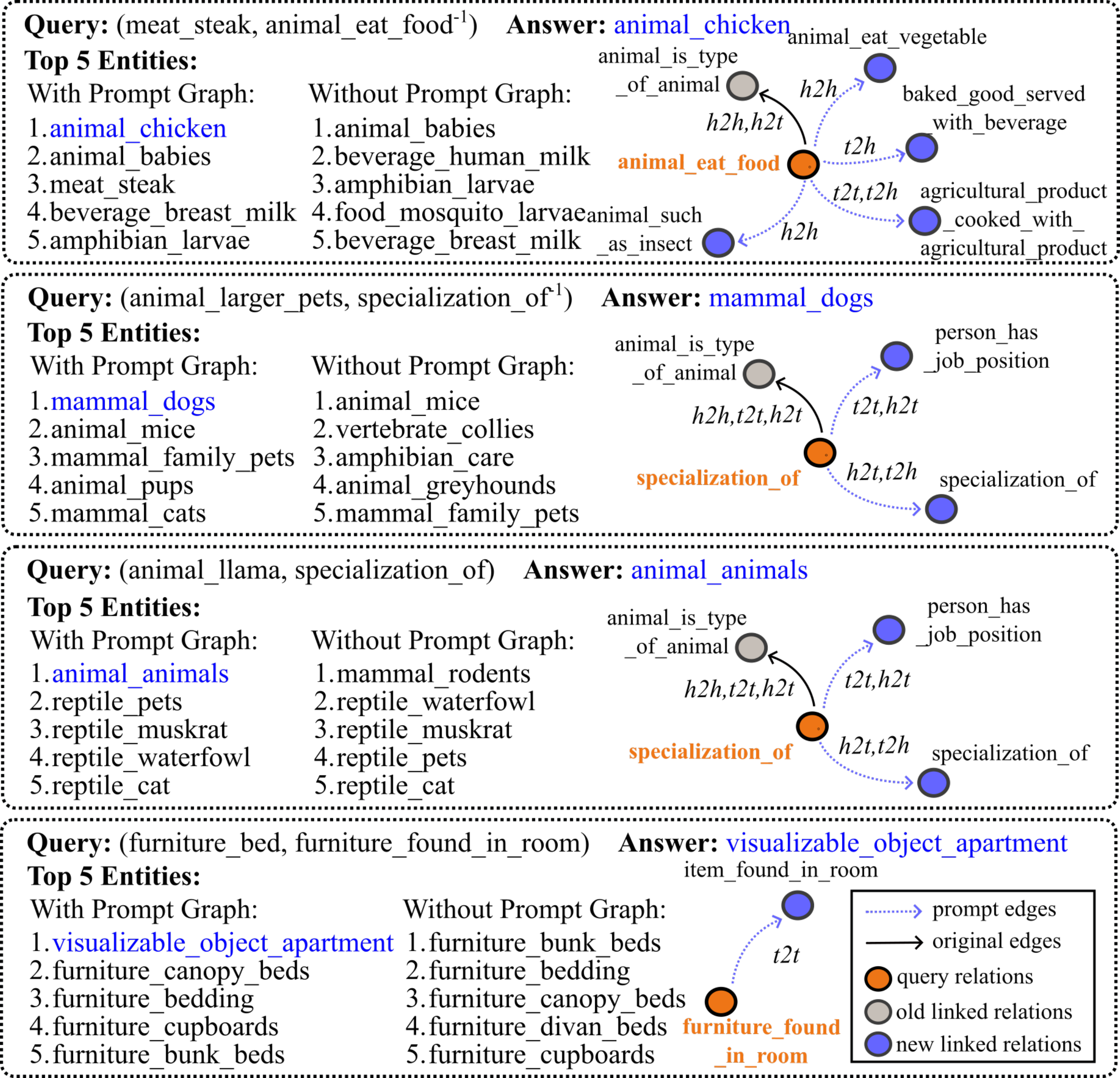}
\vspace{-3mm}
\caption{Case studies on the one-shot NL:v4 dataset.}
\label{fig:appcs}
\vspace{-5mm}
\end{figure*}

\textbf{Few-shot and Unseen Relation Reasoning:} 
To generalize to unseen relations, early efforts have explored meta-learning for few-shot link prediction, which predicts KG facts of unseen relations using a limited number of support triples \cite{FewKGR-EMNLP19,FewKGR-AAAI20,FewKGR-NIPS22,FewKGR-SIGKDD23,FewKGR-ICLR23}. 
However, those methods cannot work on an entire unseen inference graph.
Recent approaches focus on constructing graphs of relations to generalize to unseen relations \cite{rmpi,mtdea}. 
RMPI~\cite{rmpi} transforms the surrounding subgraph of a triple in the original KG into a new relation-view graph, then learns the embedding of an unseen relation from the relational subgraph. 
InGram~\cite{ingram} relies on a featurization strategy based on the discretization of node degrees, which is effective only for KGs with a similar relational distribution and falls short in transferring to arbitrary KGs.
ISDEA~\citep{isdea} utilizes relation exchangeability in multi-relational graphs, which imposes stringent assumptions on the class of transferable relations coming from the same distribution. 
ULTRA \cite{ultra} employs a pre-training and fine-tuning framework, with a global relation graph extracted from the entire set of KG triples. 
However, the effectiveness of the above methods is contingent on sufficient support triples for the unseen relations in the inference KG, struggling with few-shot relation types in low-resource scenarios that limit their applicability.

\textbf{Text-based methods via Language Models:}
In recent years, the AI community has witnessed the emergence of numerous powerful pre-trained language Models (PLMs), such as BERT~\citep{Bert}, Llama-2~\citep{Llama2}, and GPT-4~\citep{Gpt4} which are driving huge advancements and lead to the pursuit of possible Artificial General Intelligence (AGI) \cite{ouyang2022training,bubeck2023sparks}.
A line of text-based KG inductive reasoning methods like BLP~\citep{TextKGR-BLP-WWW21}, KEPLER~\citep{TextKGR-KEPLER-TACL21}, StATIK~\citep{TextKGR-StATIK-NAACL21}, RAILD~\citep{TextKGR-RAILD-IJCKG22} rely on textual descriptions of entities and relations and use pre-trained language models (PLMs) like BERT~\citep{Bert} to encode them. 
To seek higher prediction performance, current methods mostly prefer to fine-tune PLMs which requires high computational complexity and limits its generalizability to other KGs with different formats of textual descriptions.
CSProm-KG \cite{CSPromKG-ACL23} and PDKGC \cite{PDKGC-arxiv23} perform KG link prediction via a frozen PLM with only the prompts trained. 
By combining graph structure features and textual representations, these methods perform well in transductive KG reasoning tasks. 
However, the trainable prompts are tuned for a fixed relation vocabulary and cannot be generalized to inductive reasoning.
Recent work \cite{LLMKGC-23arxiv} verifies the reasoning performance of large language models by converting a query into textual names of the query entity and relation.
We consider this family of methods to be orthogonal to our work. 
Our approach operates under the low-resource assumption that the graphs lack explicit entity descriptions, focusing instead on leveraging structural graph information and relation descriptions. 


\begin{table*}[!h]
\scriptsize
\centering
\begin{tabular}{llllllll}
\hline
K-shot & Model & FB:v1 & FB:v2 & FB:v3 & FB:v4 & WK:v1 & WK:v2 \\
\hline
3shot & NBFNet & 0.084 (±0.000) & 0.016 (±0.000) & 0.012 (±0.000) & 0.001 (±0.000) & 0.011 (±0.000) & 0.002 (±0.000) \\
3shot & REDGNN & 0.041 (±0.000) & 0.035 (±0.001) & 0.056 (±0.001) & 0.025 (±0.001) & 0.001 (±0.000) & 0.003 (±0.001) \\
3shot & InGram & 0.071 (±0.006) & 0.029 (±0.002) & 0.062 (±0.007) & 0.035 (±0.005) & 0.051 (±0.013) & 0.001 (±0.001) \\
3shot & DEqInGram & 0.102 (±0.004) & 0.035 (±0.005) & 0.108 (±0.010) & 0.038 (±0.006) & 0.021 (±0.008) & 0.009 (±0.001) \\
3shot & ISDEA & 0.009 (±0.001) & 0.005 (±0.001) & 0.005 (±0.000) & 0.005 (±0.000) & 0.008 (±0.001) & 0.003 (±0.000) \\
3shot & ULTRA(3g) & 0.241 (±0.007) & 0.212 (±0.004) & 0.228 (±0.012) & 0.186 (±0.009) & 0.298 (±0.017) & 0.050 (±0.010) \\
3shot & ULTRA(4g) & 0.268 (±0.007) & 0.211 (±0.008) & 0.218 (±0.009) & 0.196 (±0.010) & 0.123 (±0.028) & 0.041 (±0.010) \\
3shot & ULTRA(50g) & 0.248 (±0.004) & 0.209 (±0.007) & 0.215 (±0.015) & 0.197 (±0.005) & 0.129 (±0.008) & 0.037 (±0.012) \\
3shot & Our(Llama2-7B) & 0.331 (±0.006) & 0.290 (±0.017) & 0.260 (±0.008) & 0.232 (±0.008) & 0.321 (±0.024) & 0.052 (±0.008) \\
3shot & Our(Llama2-13B) & 0.313 (±0.006) & 0.289 (±0.013) & 0.253 (±0.005) & 0.225 (±0.009) & 0.333 (±0.010) & 0.053 (±0.007) \\
3shot & Our(Mistral-7B) & 0.316 (±0.003) & 0.294 (±0.012) & 0.253 (±0.011) & 0.231 (±0.005) & 0.325 (±0.018) & 0.053 (±0.008) \\
3shot & Our(GPT-3.5) & 0.325 (±0.002) & 0.286 (±0.011) & 0.256 (±0.011) & 0.230 (±0.008) & 0.331 (±0.011) & 0.054 (±0.009) \\
3shot & Our(GPT-4) & 0.332 (±0.004) & 0.290 (±0.011) & 0.256 (±0.013) & 0.237 (±0.004) & 0.326 (±0.013) & 0.058 (±0.008) \\
\hline
1shot & NBFNet & 0.079 (±0.000) & 0.015 (±0.000) & 0.012 (±0.000) & 0.001 (±0.000) & 0.008 (±0.000) & 0.002 (±0.000) \\
1shot & REDGNN & 0.038 (±0.000) & 0.034 (±0.001) & 0.055 (±0.001) & 0.020 (±0.001) & 0.001 (±0.000) & 0.002 (±0.000) \\
1shot & InGram & 0.050 (±0.007) & 0.032 (±0.003) & 0.035 (±0.011) & 0.025 (±0.006) & 0.060 (±0.010) & 0.001 (±0.001) \\
1shot & DEqInGram & 0.078 (±0.011) & 0.029 (±0.004) & 0.082 (±0.009) & 0.034 (±0.004) & 0.019 (±0.003) & 0.008 (±0.002) \\
1shot & ISDEA & 0.008 (±0.001) & 0.004 (±0.001) & 0.004 (±0.001) & 0.004 (±0.000) & 0.007 (±0.002) & 0.002 (±0.001) \\
1shot & ULTRA(3g) & 0.182 (±0.011) & 0.153 (±0.004) & 0.159 (±0.009) & 0.108 (±0.010) & 0.221 (±0.020) & 0.030 (±0.001) \\
1shot & ULTRA(4g) & 0.190 (±0.008) & 0.149 (±0.008) & 0.156 (±0.012) & 0.132 (±0.013) & 0.051 (±0.009) & 0.023 (±0.003) \\
1shot & ULTRA(50g) & 0.180 (±0.010) & 0.142 (±0.005) & 0.148 (±0.016) & 0.141 (±0.011) & 0.062 (±0.029) & 0.020 (±0.002) \\
1shot & Our(Llama2-7B) & 0.280 (±0.013) & 0.250 (±0.003) & 0.228 (±0.013) & 0.185 (±0.006) & 0.259 (±0.029) & 0.031 (±0.004) \\
1shot & Our(Llama2-13B) & 0.270 (±0.008) & 0.249 (±0.010) & 0.221 (±0.010) & 0.174 (±0.011) & 0.275 (±0.027) & 0.030 (±0.004) \\
1shot & Our(Mistral-7B) & 0.281 (±0.012) & 0.251 (±0.001) & 0.224 (±0.008) & 0.179 (±0.011) & 0.267 (±0.023) & 0.035 (±0.004) \\
1shot & Our(GPT-3.5) & 0.273 (±0.008) & 0.246 (±0.009) & 0.228 (±0.004) & 0.190 (±0.007) & 0.259 (±0.023) & 0.034 (±0.005) \\
1shot & Our(GPT-4) & 0.281 (±0.011) & 0.248 (±0.009) & 0.226 (±0.005) & 0.194 (±0.008) & 0.262 (±0.029) & 0.035 (±0.006) \\
\hline
0shot & NBFNet & 0.079 (±0.000) & 0.015 (±0.000) & 0.012 (±0.000) & 0.001 (±0.000) & 0.008 (±0.000) & 0.002 (±0.000) \\
0shot & REDGNN & 0.038 (±0.000) & 0.033 (±0.000) & 0.054 (±0.000) & 0.020 (±0.000) & 0.001 (±0.000) & 0.002 (±0.000) \\
0shot & InGram & 0.009 (±0.000) & 0.021 (±0.000) & 0.025 (±0.000) & 0.014 (±0.000) & 0.041 (±0.000) & 0.001 (±0.000) \\
0shot & DEqInGram & 0.015 (±0.001) & 0.020 (±0.001) & 0.037 (±0.005) & 0.024 (±0.003) & 0.053 (±0.029) & 0.009 (±0.001) \\
0shot & ISDEA & 0.007 (±0.000) & 0.001 (±0.000) & 0.002 (±0.000) & 0.003 (±0.000) & 0.003 (±0.001) & 0.001 (±0.001) \\
0shot & ULTRA(3g) & 0.048 (±0.000) & 0.035 (±0.000) & 0.034 (±0.000) & 0.028 (±0.000) & 0.019 (±0.000) & 0.006 (±0.000) \\
0shot & ULTRA(4g) & 0.053 (±0.000) & 0.041 (±0.000) & 0.033 (±0.000) & 0.030 (±0.000) & 0.020 (±0.000) & 0.006 (±0.000) \\
0shot & ULTRA(50g) & 0.035 (±0.000) & 0.024 (±0.000) & 0.023 (±0.000) & 0.029 (±0.000) & 0.019 (±0.000) & 0.006 (±0.000) \\
0shot & Our(Llama2-7B) & 0.092 (±0.000) & 0.099 (±0.000) & 0.064 (±0.000) & 0.067 (±0.000) & 0.016 (±0.000) & 0.008 (±0.000) \\
0shot & Our(Llama2-13B) & 0.110 (±0.000) & 0.097 (±0.000) & 0.067 (±0.000) & 0.059 (±0.000) & 0.036 (±0.000) & 0.008 (±0.000) \\
0shot & Our(Mistral-7B) & 0.152 (±0.000) & 0.139 (±0.000) & 0.131 (±0.000) & 0.097 (±0.000) & 0.018 (±0.000) & 0.007 (±0.000) \\
0shot & Our(GPT-3.5) & 0.181 (±0.000) & 0.166 (±0.000) & 0.134 (±0.000) & 0.120 (±0.000) & 0.191 (±0.000) & 0.011 (±0.000) \\
0shot & Our(GPT-4) & 0.186 (±0.000) & 0.168 (±0.000) & 0.141 (±0.000) & 0.108 (±0.000) & 0.235 (±0.000) & 0.013 (±0.000)\\
\hline
\end{tabular}
\caption{Detailed inductive reasoning results (1), evaluated with MRR.}
\label{table:allMRR1}
\end{table*}

\begin{table*}[]
\scriptsize
\centering
\begin{tabular}{llllllll}
\hline
K-shot & Model & WK:v3 & WK:v4 & NL:v1 & NL:v2 & NL:v3 & NL:v4 \\
\hline
3shot & NBFNet & 0.005 (±0.000) & 0.002 (±0.000) & 0.003 (±0.000) & 0.019 (±0.000) & 0.009 (±0.000) & 0.002 (±0.000) \\
3shot & REDGNN & 0.001 (±0.000) & 0.001 (±0.000) & 0.063 (±0.000) & 0.106 (±0.000) & 0.092 (±0.003) & 0.014 (±0.000) \\
3shot & InGram & 0.015 (±0.008) & 0.016 (±0.005) & 0.060 (±0.007) & 0.071 (±0.014) & 0.063 (±0.003) & 0.075 (±0.009) \\
3shot & DEqInGram & 0.012 (±0.007) & 0.068 (±0.007) & 0.120 (±0.014) & 0.083 (±0.014) & 0.103 (±0.010) & 0.133 (±0.017) \\
3shot & ISDEA & 0.008 (±0.001) & 0.002 (±0.000) & 0.019 (±0.005) & 0.017 (±0.003) & 0.017 (±0.005) & 0.009 (±0.000) \\
3shot & ULTRA(3g) & 0.135 (±0.018) & 0.078 (±0.013) & 0.197 (±0.028) & 0.211 (±0.033) & 0.209 (±0.010) & 0.245 (±0.027) \\
3shot & ULTRA(4g) & 0.137 (±0.020) & 0.072 (±0.014) & 0.192 (±0.022) & 0.192 (±0.033) & 0.181 (±0.021) & 0.216 (±0.025) \\
3shot & ULTRA(50g) & 0.139 (±0.023) & 0.068 (±0.011) & 0.169 (±0.029) & 0.180 (±0.040) & 0.150 (±0.031) & 0.184 (±0.028) \\
3shot & Our(Llama2-7B) & 0.137 (±0.024) & 0.076 (±0.012) & 0.225 (±0.023) & 0.243 (±0.020) & 0.215 (±0.011) & 0.261 (±0.019) \\
3shot & Our(Llama2-13B) & 0.137 (±0.022) & 0.076 (±0.013) & 0.217 (±0.021) & 0.232 (±0.014) & 0.223 (±0.013) & 0.264 (±0.026) \\
3shot & Our(Mistral-7B) & 0.139 (±0.025) & 0.076 (±0.012) & 0.223 (±0.027) & 0.222 (±0.041) & 0.229 (±0.017) & 0.261 (±0.021) \\
3shot & Our(GPT-3.5) & 0.140 (±0.023) & 0.078 (±0.013) & 0.219 (±0.012) & 0.235 (±0.012) & 0.233 (±0.014) & 0.261 (±0.021) \\
3shot & Our(GPT-4) & 0.141 (±0.020) & 0.077 (±0.013) & 0.229 (±0.021) & 0.239 (±0.020) & 0.230 (±0.012) & 0.249 (±0.025) \\
\hline
1shot & NBFNet & 0.002 (±0.000) & 0.002 (±0.000) & 0.004 (±0.000) & 0.076 (±0.000) & 0.007 (±0.000) & 0.002 (±0.000) \\
1shot & REDGNN & 0.001 (±0.000) & 0.001 (±0.000) & 0.047 (±0.001) & 0.102 (±0.000) & 0.071 (±0.002) & 0.015 (±0.000) \\
1shot & InGram & 0.006 (±0.002) & 0.007 (±0.001) & 0.060 (±0.006) & 0.050 (±0.002) & 0.047 (±0.002) & 0.062 (±0.002) \\
1shot & DEqInGram & 0.012 (±0.002) & 0.019 (±0.003) & 0.135 (±0.028) & 0.073 (±0.022) & 0.084 (±0.004) & 0.120 (±0.012) \\
1shot & ISDEA & 0.008 (±0.002) & 0.001 (±0.000) & 0.020 (±0.003) & 0.015 (±0.004) & 0.016 (±0.005) & 0.009 (±0.000) \\
1shot & ULTRA(3g) & 0.100 (±0.018) & 0.023 (±0.001) & 0.150 (±0.018) & 0.155 (±0.035) & 0.153 (±0.016) & 0.199 (±0.017) \\
1shot & ULTRA(4g) & 0.100 (±0.020) & 0.020 (±0.002) & 0.160 (±0.012) & 0.137 (±0.031) & 0.100 (±0.011) & 0.118 (±0.011) \\
1shot & ULTRA(50g) & 0.099 (±0.023) & 0.020 (±0.002) & 0.137 (±0.014) & 0.113 (±0.009) & 0.093 (±0.003) & 0.140 (±0.022) \\
1shot & Our(Llama2-7B) & 0.103 (±0.024) & 0.023 (±0.003) & 0.207 (±0.030) & 0.189 (±0.016) & 0.174 (±0.004) & 0.222 (±0.027) \\
1shot & Our(Llama2-13B) & 0.100 (±0.024) & 0.026 (±0.002) & 0.196 (±0.005) & 0.192 (±0.022) & 0.176 (±0.003) & 0.215 (±0.015) \\
1shot & Our(Mistral-7B) & 0.104 (±0.025) & 0.023 (±0.001) & 0.193 (±0.014) & 0.184 (±0.037) & 0.171 (±0.003) & 0.206 (±0.028) \\
1shot & Our(GPT-3.5) & 0.103 (±0.023) & 0.025 (±0.003) & 0.196 (±0.017) & 0.193 (±0.018) & 0.169 (±0.005) & 0.223 (±0.014) \\
1shot & Our(GPT-4) & 0.104 (±0.023) & 0.024 (±0.002) & 0.203 (±0.021) & 0.193 (±0.042) & 0.171 (±0.004) & 0.200 (±0.028) \\
\hline
0shot & NBFNet & 0.002 (±0.000) & 0.002 (±0.000) & 0.004 (±0.000) & 0.076 (±0.000) & 0.007 (±0.000) & 0.002 (±0.000) \\
0shot & REDGNN & 0.001 (±0.000) & 0.001 (±0.000) & 0.046 (±0.000) & 0.102 (±0.000) & 0.062 (±0.000) & 0.015 (±0.000) \\
0shot & InGram & 0.001 (±0.000) & 0.001 (±0.000) & 0.032 (±0.000) & 0.042 (±0.000) & 0.015 (±0.000) & 0.055 (±0.000) \\
0shot & DEqInGram & 0.004 (±0.001) & 0.009 (±0.001) & 0.083 (±0.007) & 0.067 (±0.006) & 0.055 (±0.006) & 0.049 (±0.001) \\
0shot & ISDEA & 0.003 (±0.000) & 0.000 (±0.000) & 0.006 (±0.000) & 0.008 (±0.001) & 0.003 (±0.000) & 0.006 (±0.002)\\
0shot & ULTRA(3g) & 0.013 (±0.000) & 0.004 (±0.000) & 0.037 (±0.000) & 0.021 (±0.000) & 0.024 (±0.000) & 0.022 (±0.000) \\
0shot & ULTRA(4g) & 0.013 (±0.000) & 0.003 (±0.000) & 0.034 (±0.000) & 0.024 (±0.000) & 0.024 (±0.000) & 0.022 (±0.000) \\
0shot & ULTRA(50g) & 0.013 (±0.000) & 0.003 (±0.000) & 0.033 (±0.000) & 0.027 (±0.000) & 0.024 (±0.000) & 0.017 (±0.000) \\
0shot & Our(Llama2-7B) & 0.013 (±0.000) & 0.006 (±0.000) & 0.103 (±0.000) & 0.125 (±0.000) & 0.058 (±0.000) & 0.102 (±0.000) \\
0shot & Our(Llama2-13B) & 0.015 (±0.000) & 0.011 (±0.000) & 0.113 (±0.000) & 0.128 (±0.000) & 0.094 (±0.000) & 0.100 (±0.000) \\
0shot & Our(Mistral-7B) & 0.014 (±0.000) & 0.009 (±0.000) & 0.101 (±0.000) & 0.082 (±0.000) & 0.082 (±0.000) & 0.067 (±0.000) \\
0shot & Our(GPT-3.5) & 0.017 (±0.000) & 0.009 (±0.000) & 0.114 (±0.000) & 0.125 (±0.000) & 0.085 (±0.000) & 0.110 (±0.000) \\
0shot & Our(GPT-4) & 0.017 (±0.000) & 0.009 (±0.000) & 0.091 (±0.000) & 0.095 (±0.000) & 0.089 (±0.000) & 0.116 (±0.000)\\
\hline
\end{tabular}
\caption{Detailed inductive reasoning results (2), evaluated with MRR.}
\label{table:allMRR2}
\end{table*}

\begin{table*}[]
\scriptsize
\centering
\begin{tabular}{llllllll}
\hline
K-shot & Model & FB:v1 & FB:v2 & FB:v3 & FB:v4 & WK:v1 & WK:v2 \\
\hline
3shot & NBFNet & 0.092 (±0.000) & 0.019 (±0.000) & 0.013 (±0.000) & 0.000 (±0.000) & 0.033 (±0.000) & 0.001 (±0.000) \\
3shot & REDGNN & 0.068 (±0.001) & 0.066 (±0.001) & 0.110 (±0.001) & 0.065 (±0.001) & 0.000 (±0.000) & 0.003 (±0.002) \\
3shot & InGram & 0.098 (±0.007) & 0.062 (±0.005) & 0.114 (±0.004) & 0.070 (±0.007) & 0.070 (±0.008) & 0.002 (±0.001) \\
3shot & DEqInGram & 0.076 (±0.006) & 0.009 (±0.002) & 0.070 (±0.012) & 0.011 (±0.004) & 0.006 (±0.005) & 0.002 (±0.001) \\
3shot & ISDEA & 0.018 (±0.001) & 0.013 (±0.000) & 0.009 (±0.001) & 0.008 (±0.001) & 0.014 (±0.002) & 0.003 (±0.001) \\
3shot & ULTRA(3g) & 0.359 (±0.008) & 0.334 (±0.009) & 0.335 (±0.015) & 0.319 (±0.017) & 0.430 (±0.004) & 0.092 (±0.031) \\
3shot & ULTRA(4g) & 0.404 (±0.023) & 0.358 (±0.018) & 0.327 (±0.011) & 0.316 (±0.019) & 0.138 (±0.031) & 0.063 (±0.025) \\
3shot & ULTRA(50g) & 0.368 (±0.016) & 0.354 (±0.008) & 0.329 (±0.015) & 0.310 (±0.015) & 0.157 (±0.010) & 0.055 (±0.026) \\
3shot & Our(Llama2-7B) & 0.513 (±0.012) & 0.480 (±0.014) & 0.408 (±0.017) & 0.377 (±0.023) & 0.461 (±0.016) & 0.108 (±0.021) \\
3shot & Our(Llama2-13B) & 0.492 (±0.008) & 0.468 (±0.015) & 0.405 (±0.019) & 0.358 (±0.022) & 0.463 (±0.021) & 0.116 (±0.020) \\
3shot & Our(Mistral-7B) & 0.494 (±0.005) & 0.467 (±0.017) & 0.402 (±0.019) & 0.367 (±0.023) & 0.461 (±0.007) & 0.115 (±0.020) \\
3shot & Our(GPT-3.5) & 0.505 (±0.005) & 0.471 (±0.014) & 0.408 (±0.016) & 0.380 (±0.019) & 0.455 (±0.002) & 0.115 (±0.018) \\
3shot & Our(GPT-4) & 0.517 (±0.012) & 0.473 (±0.014) & 0.404 (±0.014) & 0.375 (±0.023) & 0.456 (±0.001) & 0.118 (±0.028) \\
\hline
1shot & NBFNet & 0.087 (±0.000) & 0.018 (±0.000) & 0.013 (±0.000) & 0.000 (±0.000) & 0.019 (±0.000) & 0.001 (±0.000) \\
1shot & REDGNN & 0.063 (±0.000) & 0.064 (±0.000) & 0.103 (±0.000) & 0.054 (±0.001) & 0.002 (±0.000) & 0.001 (±0.000) \\
1shot & InGram & 0.069 (±0.008) & 0.057 (±0.003) & 0.062 (±0.022) & 0.049 (±0.010) & 0.080 (±0.023) & 0.003 (±0.002) \\
1shot & DEqInGram & 0.063 (±0.011) & 0.007 (±0.002) & 0.050 (±0.010) & 0.013 (±0.003) & 0.004 (±0.002) & 0.002 (±0.001) \\
1shot & ISDEA & 0.022 (±0.002) & 0.008 (±0.001) & 0.007 (±0.001) & 0.006 (±0.000) & 0.011 (±0.004) & 0.002 (±0.001) \\
1shot & ULTRA(3g) & 0.270 (±0.010) & 0.256 (±0.010) & 0.235 (±0.014) & 0.204 (±0.020) & 0.249 (±0.031) & 0.042 (±0.003) \\
1shot & ULTRA(4g) & 0.296 (±0.011) & 0.273 (±0.016) & 0.230 (±0.013) & 0.221 (±0.011) & 0.043 (±0.020) & 0.025 (±0.002) \\
1shot & ULTRA(50g) & 0.275 (±0.009) & 0.254 (±0.012) & 0.230 (±0.025) & 0.235 (±0.013) & 0.065 (±0.035) & 0.023 (±0.004) \\
1shot & Our(Llama2-7B) & 0.442 (±0.022) & 0.417 (±0.013) & 0.363 (±0.023) & 0.298 (±0.005) & 0.371 (±0.033) & 0.056 (±0.017) \\
1shot & Our(Llama2-13B) & 0.445 (±0.011) & 0.404 (±0.010) & 0.350 (±0.014) & 0.280 (±0.021) & 0.384 (±0.037) & 0.054 (±0.012) \\
1shot & Our(Mistral-7B) & 0.441 (±0.013) & 0.404 (±0.004) & 0.359 (±0.006) & 0.295 (±0.013) & 0.381 (±0.029) & 0.061 (±0.019) \\
1shot & Our(GPT-3.5) & 0.438 (±0.022) & 0.417 (±0.011) & 0.369 (±0.005) & 0.324 (±0.007) & 0.381 (±0.029) & 0.058 (±0.017) \\
1shot & Our(GPT-4) & 0.451 (±0.012) & 0.409 (±0.018) & 0.370 (±0.007) & 0.319 (±0.006) & 0.377 (±0.040) & 0.063 (±0.021) \\
\hline
0shot & NBFNet & 0.087 (±0.000) & 0.018 (±0.000) & 0.013 (±0.000) & 0.000 (±0.000) & 0.019 (±0.000) & 0.001 (±0.000) \\
0shot & REDGNN & 0.063 (±0.000) & 0.064 (±0.000) & 0.103 (±0.000) & 0.054 (±0.000) & 0.002 (±0.000) & 0.001 (±0.000) \\
0shot & InGram & 0.017 (±0.000) & 0.043 (±0.000) & 0.058 (±0.000) & 0.027 (±0.000) & 0.054 (±0.000) & 0.001 (±0.000) \\
0shot & DEqInGram & 0.003 (±0.002) & 0.004 (±0.001) & 0.013 (±0.005) & 0.007 (±0.001) & 0.023 (±0.031) & 0.001 (±0.001) \\
0shot & ISDEA & 0.017 (±0.000) & 0.001 (±0.000) & 0.002 (±0.000) & 0.002 (±0.000) & 0.002 (±0.002) & 0.000 (±0.001) \\
0shot & ULTRA(3g) & 0.145 (±0.000) & 0.128 (±0.000) & 0.101 (±0.000) & 0.098 (±0.000) & 0.019 (±0.000) & 0.008 (±0.000) \\
0shot & ULTRA(4g) & 0.155 (±0.000) & 0.148 (±0.000) & 0.101 (±0.000) & 0.107 (±0.000) & 0.018 (±0.000) & 0.010 (±0.000) \\
0shot & ULTRA(50g) & 0.103 (±0.000) & 0.100 (±0.000) & 0.063 (±0.000) & 0.100 (±0.000) & 0.023 (±0.000) & 0.012 (±0.000) \\
0shot & Our(Llama2-7B) & 0.193 (±0.000) & 0.203 (±0.000) & 0.133 (±0.000) & 0.153 (±0.000) & 0.029 (±0.000) & 0.010 (±0.000) \\
0shot & Our(Llama2-13B) & 0.209 (±0.000) & 0.179 (±0.000) & 0.136 (±0.000) & 0.121 (±0.000) & 0.078 (±0.000) & 0.011 (±0.000) \\
0shot & Our(Mistral-7B) & 0.254 (±0.000) & 0.254 (±0.000) & 0.222 (±0.000) & 0.195 (±0.000) & 0.043 (±0.000) & 0.012 (±0.000) \\
0shot & Our(GPT-3.5) & 0.300 (±0.000) & 0.267 (±0.000) & 0.240 (±0.000) & 0.225 (±0.000) & 0.304 (±0.000) & 0.020 (±0.000) \\
0shot & Our(GPT-4) & 0.284 (±0.000) & 0.251 (±0.000) & 0.256 (±0.000) & 0.213 (±0.000) & 0.350 (±0.000) & 0.024 (±0.000)\\
\hline
\end{tabular}
\caption{Detailed inductive reasoning results (1), evaluated with Hits@10.}
\label{table:allH101}
\end{table*}

\begin{table*}[]
\scriptsize
\centering
\begin{tabular}{llllllll}
\hline
K-shot & Model & WK:v3 & WK:v4 & NL:v1 & NL:v2 & NL:v3 & NL:v4 \\
\hline
3shot & NBFNet & 0.000 (±0.000) & 0.000 (±0.000) & 0.000 (±0.000) & 0.014 (±0.000) & 0.005 (±0.000) & 0.000 (±0.000) \\
3shot & REDGNN & 0.000 (±0.000) & 0.000 (±0.000) & 0.081 (±0.000) & 0.171 (±0.000) & 0.136 (±0.007) & 0.025 (±0.000) \\
3shot & InGram & 0.022 (±0.011) & 0.023 (±0.004) & 0.122 (±0.029) & 0.124 (±0.007) & 0.138 (±0.014) & 0.149 (±0.014) \\
3shot & DEqInGram & 0.005 (±0.003) & 0.052 (±0.006) & 0.068 (±0.009) & 0.046 (±0.017) & 0.043 (±0.007) & 0.074 (±0.013) \\
3shot & ISDEA & 0.012 (±0.004) & 0.003 (±0.000) & 0.035 (±0.009) & 0.031 (±0.006) & 0.035 (±0.004) & 0.019 (±0.001) \\
3shot & ULTRA(3g) & 0.172 (±0.031) & 0.114 (±0.007) & 0.275 (±0.058) & 0.291 (±0.048) & 0.324 (±0.046) & 0.369 (±0.018) \\
3shot & ULTRA(4g) & 0.176 (±0.032) & 0.092 (±0.010) & 0.258 (±0.043) & 0.255 (±0.020) & 0.265 (±0.027) & 0.300 (±0.017) \\
3shot & ULTRA(50g) & 0.166 (±0.033) & 0.086 (±0.009) & 0.227 (±0.040) & 0.220 (±0.037) & 0.219 (±0.041) & 0.262 (±0.025) \\
3shot & Our(Llama2-7B) & 0.186 (±0.029) & 0.111 (±0.007) & 0.335 (±0.063) & 0.372 (±0.031) & 0.335 (±0.041) & 0.394 (±0.001) \\
3shot & Our(Llama2-13B) & 0.186 (±0.032) & 0.110 (±0.008) & 0.320 (±0.056) & 0.367 (±0.029) & 0.354 (±0.042) & 0.403 (±0.017) \\
3shot & Our(Mistral-7B) & 0.188 (±0.030) & 0.108 (±0.006) & 0.323 (±0.049) & 0.333 (±0.062) & 0.354 (±0.030) & 0.394 (±0.007) \\
3shot & Our(GPT-3.5) & 0.187 (±0.032) & 0.113 (±0.009) & 0.342 (±0.027) & 0.351 (±0.066) & 0.355 (±0.049) & 0.411 (±0.001) \\
3shot & Our(GPT-4) & 0.188 (±0.036) & 0.112 (±0.011) & 0.330 (±0.062) & 0.352 (±0.045) & 0.338 (±0.060) & 0.398 (±0.017) \\
\hline
1shot & NBFNet & 0.000 (±0.000) & 0.000 (±0.000) & 0.000 (±0.000) & 0.071 (±0.000) & 0.004 (±0.000) & 0.000 (±0.000) \\
1shot & REDGNN & 0.004 (±0.000) & 0.000 (±0.000) & 0.063 (±0.003) & 0.164 (±0.000) & 0.099 (±0.004) & 0.029 (±0.000) \\
1shot & InGram & 0.014 (±0.005) & 0.009 (±0.001) & 0.096 (±0.013) & 0.082 (±0.006) & 0.085 (±0.007) & 0.116 (±0.015) \\
1shot & DEqInGram & 0.004 (±0.004) & 0.012 (±0.003) & 0.096 (±0.031) & 0.043 (±0.024) & 0.035 (±0.005) & 0.066 (±0.005) \\
1shot & ISDEA & 0.011 (±0.002) & 0.001 (±0.000) & 0.026 (±0.005) & 0.027 (±0.007) & 0.023 (±0.012) & 0.018 (±0.002) \\
1shot & ULTRA(3g) & 0.101 (±0.026) & 0.029 (±0.003) & 0.216 (±0.031) & 0.194 (±0.033) & 0.223 (±0.016) & 0.255 (±0.024) \\
1shot & ULTRA(4g) & 0.101 (±0.026) & 0.020 (±0.002) & 0.183 (±0.010) & 0.171 (±0.036) & 0.151 (±0.003) & 0.160 (±0.012) \\
1shot & ULTRA(50g) & 0.096 (±0.025) & 0.020 (±0.002) & 0.185 (±0.032) & 0.176 (±0.033) & 0.155 (±0.003) & 0.186 (±0.018) \\
1shot & Our(Llama2-7B) & 0.116 (±0.030) & 0.029 (±0.004) & 0.295 (±0.022) & 0.277 (±0.033) & 0.247 (±0.014) & 0.311 (±0.028) \\
1shot & Our(Llama2-13B) & 0.119 (±0.030) & 0.034 (±0.002) & 0.272 (±0.011) & 0.279 (±0.031) & 0.256 (±0.009) & 0.312 (±0.015) \\
1shot & Our(Mistral-7B) & 0.119 (±0.032) & 0.031 (±0.001) & 0.270 (±0.012) & 0.259 (±0.049) & 0.268 (±0.009) & 0.285 (±0.027) \\
1shot & Our(GPT-3.5) & 0.118 (±0.029) & 0.035 (±0.003) & 0.292 (±0.029) & 0.291 (±0.050) & 0.266 (±0.018) & 0.322 (±0.025) \\
1shot & Our(GPT-4) & 0.118 (±0.022) & 0.035 (±0.004) & 0.297 (±0.033) & 0.267 (±0.026) & 0.260 (±0.009) & 0.300 (±0.015) \\
\hline
0shot & NBFNet & 0.000 (±0.000) & 0.000 (±0.000) & 0.000 (±0.000) & 0.071 (±0.000) & 0.004 (±0.000) & 0.000 (±0.000) \\
0shot & REDGNN & 0.000 (±0.000) & 0.000 (±0.000) & 0.060 (±0.000) & 0.164 (±0.000) & 0.088 (±0.000) & 0.029 (±0.000) \\
0shot & InGram & 0.002 (±0.000) & 0.002 (±0.000) & 0.070 (±0.000) & 0.070 (±0.000) & 0.029 (±0.000) & 0.081 (±0.000) \\
0shot & DEqInGram & 0.000 (±0.001) & 0.001 (±0.001) & 0.040 (±0.018) & 0.032 (±0.002) & 0.022 (±0.005) & 0.020 (±0.001) \\
0shot & ISDEA & 0.004 (±0.001) & 0.000 (±0.000) & 0.009 (±0.003) & 0.014 (±0.004) & 0.003 (±0.002) & 0.012 (±0.005) \\
0shot & ULTRA(3g) & 0.013 (±0.000) & 0.004 (±0.000) & 0.113 (±0.000) & 0.068 (±0.000) & 0.061 (±0.000) & 0.060 (±0.000) \\
0shot & ULTRA(4g) & 0.013 (±0.000) & 0.004 (±0.000) & 0.096 (±0.000) & 0.071 (±0.000) & 0.059 (±0.000) & 0.061 (±0.000) \\
0shot & ULTRA(50g) & 0.013 (±0.000) & 0.004 (±0.000) & 0.089 (±0.000) & 0.077 (±0.000) & 0.058 (±0.000) & 0.046 (±0.000) \\
0shot & Our(Llama2-7B) & 0.016 (±0.000) & 0.007 (±0.000) & 0.180 (±0.000) & 0.190 (±0.000) & 0.095 (±0.000) & 0.161 (±0.000) \\
0shot & Our(Llama2-13B) & 0.016 (±0.000) & 0.016 (±0.000) & 0.191 (±0.000) & 0.184 (±0.000) & 0.155 (±0.000) & 0.148 (±0.000) \\
0shot & Our(Mistral-7B) & 0.017 (±0.000) & 0.011 (±0.000) & 0.172 (±0.000) & 0.156 (±0.000) & 0.143 (±0.000) & 0.119 (±0.000) \\
0shot & Our(GPT-3.5) & 0.030 (±0.000) & 0.014 (±0.000) & 0.178 (±0.000) & 0.178 (±0.000) & 0.153 (±0.000) & 0.177 (±0.000) \\
0shot & Our(GPT-4) & 0.031 (±0.000) & 0.018 (±0.000) & 0.183 (±0.000) & 0.140 (±0.000) & 0.147 (±0.000) & 0.191 (±0.000)\\
\hline
\end{tabular}
\caption{Detailed inductive reasoning results (2), evaluated with Hits@10.}
\label{table:allH102}
\end{table*}

\end{document}